\documentclass[10pt,conference]{IEEEtran}
\IEEEoverridecommandlockouts
\usepackage{cite}
\usepackage{amsmath,amssymb,amsfonts}
\usepackage{algorithmic}
\usepackage{graphicx}
\usepackage{textcomp}
\usepackage{booktabs}
\usepackage{array}
\usepackage{booktabs}
\usepackage{multirow}
\usepackage[table,xcdraw]{xcolor}
\usepackage{subcaption}
\usepackage{hyperref}

\def\BibTeX{{\rm B\kern-.05em{\sc i\kern-.025em b}\kern-.08em
    T\kern-.1667em\lower.7ex\hbox{E}\kern-.125emX}}
\begin{document}

\title{TaskVAE: Task-Specific Variational Autoencoders for Exemplar Generation in Continual Learning for Human Activity Recognition}
\author{\author{
    \IEEEauthorblockN{Bonpagna Kann\IEEEauthorrefmark{1}, Sandra Castellanos-Paez\IEEEauthorrefmark{3}, Romain Rombourg\IEEEauthorrefmark{2}, Philippe Lalanda\IEEEauthorrefmark{1}}
    \IEEEauthorblockA{\IEEEauthorrefmark{1}Univ. Grenoble Alpes, CNRS, Grenoble INP\footnote{Institute of Engineering Univ. Grenoble Alpes}, LIG, 38000 Grenoble, France}
    \IEEEauthorblockA{\IEEEauthorrefmark{3}School of Computing and Engineering, Univ. of Huddersfield, HD13DH, Huddersfield, United Kingdom}
    \IEEEauthorblockA{\IEEEauthorrefmark{2}Univ. Grenoble Alpes, CNRS, Grenoble INP, G2Elab, 38000 Grenoble, France
    \\firstname.lastname@univ-grenoble-alpes.fr}
}}

\maketitle

\begin{abstract}

As machine learning based systems become more integrated into daily life, they unlock new opportunities but face the challenge of adapting to dynamic data environments. Various forms of data shift—gradual, abrupt, or cyclic—threaten model accuracy, making continual adaptation essential. Continual Learning (CL) has emerged as a solution for enabling models to learn from evolving data streams while minimizing forgetting of prior knowledge. Among CL strategies, replay-based methods have proven effective, but their success relies on balancing memory constraints and retaining old class accuracy while learning new classes. 
This paper presents TaskVAE, a framework for replay-based CL in class-incremental settings. TaskVAE employs task-specific Variational Autoencoders (VAEs) to generate synthetic exemplars from previous tasks, which are then used to train the classifier alongside new task data. In contrast to traditional methods that require prior knowledge of the total class count or rely on a single VAE for all tasks, TaskVAE adapts flexibly to increasing tasks without such constraints. We focus on the Human Activity Recognition (HAR) domain, where motion patterns are analyzed using devices with IMU sensors. Unlike previous HAR studies that aggregate data across all users, our approach focuses on individual user data, better reflecting real-world scenarios where a person progressively learns new activities. Extensive experiments on 5 different HAR datasets show that TaskVAE outperforms experience replay methods, particularly with limited data, and exhibits robust performance as dataset size increases. Additionally, TaskVAE’s memory footprint is minimal, being equivalent to only 60 samples per task, while still being able to  generate an unlimited number of synthetic samples. The framework’s contributions lie in balancing memory constraints, task-specific generation, and long-term stability, making it a reliable solution for real-world applications in domains like HAR.
\end{abstract}

\section{Introduction}


Over the past few decades, the integration of pervasive computing with advanced machine learning (ML) has sparked a digital revolution, transforming sectors like industrial automation (Industry 4.0), healthcare, transportation, and smart homes. These innovations improve efficiency and change how businesses and individuals interact with technology. However, they also pose challenges, particularly in ensuring that ML-based systems can adapt to dynamic and changing data environments—an inherent characteristic of pervasive computing—where data shifts can impact model accuracy and relevance \cite{ccf, xware}.

Continual Learning (CL) stands out as a key approach, allowing models to update knowledge incrementally while retaining prior insights. CL strategies include architecture-based methods, which adjust network structures to minimize task interference but face scalability challenges; regularization techniques, which constrain weight updates to protect past knowledge but struggle with class incremental scenarios; and replay approaches, which use past data—either real or synthetic— to improve adaptability. However, most existing implementations focus on low-dimensional data and engineered features, limiting their effectiveness with the high-dimensional raw sensor data central to pervasive systems.

In this work, we propose a task-based VAE framework which trains lightweight VAEs directly on high-dimensional raw data, eliminating the need for manual feature engineering. Each VAE is used to generate synthetic samples of past classes, enabling the CL classifier to retain knowledge while adapting to new activities. By learning an approximate distribution of the data for each task, VAEs can capture important class-specific variability, facilitating the generation of representative samples, even in the presence of high-variance data.

We focus on the Human Activity Recognition (HAR) use case, where activities are represented as sequences of motion patterns with temporal correlations \cite{stisen2015smart, reiss2012introducing, Malekzadeh2019}. Using devices such as smartphones equipped with sensors (accelerometers, gyroscopes), HAR predominantly relies on machine learning to process and interpret sensor data \cite{gupta2021deep}. However, these systems inherently face the challenge of Continual Learning, as activities, subjects, and devices evolve over time \cite{usmanova2021distillation}. To compare our approach against other replay-based methods, we designed class incremental learning scenarios varying both the number of tasks and the number of classes per task, simulating real-world conditions where new activities emerge progressively. Additionally, we also conduct experiments using individual participant data, simulating real-world settings where a person performs new activities over time. Our results show that TaskVAE achieves superior generalization, especially in small datasets where other methods falter, ensuring higher accuracy in class-incremental learning. Moreover, it maintains strong stability as dataset sizes increase while remaining highly memory-efficient, generating unlimited synthetic samples with equivalent memory size to the replay-based methods.

The paper is structured as follows: Section II provides some background about HAR and CL. Section III details our VAE-based CL framework. Section IV provides the experimental details and is followed by Section V which discusses results. Section VI concludes the paper and presents  future research directions.

\section{Background and Related Work}
\label{sec:background}

In Human Activity Recognition (HAR), previous works have focused on sensor-based activity recognition, emphasizing feature extraction and training methods. Bulling et al. \cite{bulling2014tutorial} explored feature extraction techniques using statistical analysis on the features of the HAR data. Despite promising results, the extracted features were carefully engineered and heuristic, lacking systematic approach for accurate activity classification. To address this, Hammerla et al. \cite{hammerla2016deep} investigated the use of CNNs and RNNs across multiple HAR datasets \cite{chavarriaga2013opportunity, reiss2012introducing}, showing that deep learning models excel at identifying local patterns in sensor data. However, these models struggled with adapting to new, daily activities \cite{ros2013online}.

Continual Learning (CL) is gaining interest for its ability to help models adapt to new data while retaining prior knowledge, with applications in smart homes \cite{chua2022incremental}, sports training \cite{minhas2011incremental}, and healthcare \cite{sun2023few}. CL uses strategies to preserve knowledge when facing new tasks. Architecture-based methods \cite{chen2015net2net,rusu2016progressive, rakaraddi2022reinforced} adjust network structures to minimize task interference but increase complexity and scalability issues. Regularization strategies \cite{li2017learning, kemker2018measuring, aljundi2018memory} maintain a fixed architecture, constraining weight updates to protect old knowledge, but struggle in class-incremental scenarios with similar classes \cite{he2020incremental, kann2023evaluation}. 

Replay approaches \cite{rebuffi2017icarl, lopez2017gradient, ostapenko2019learning} tackle this by incorporating past task data, either real or synthetic, alongside current task data during training. While real data ensures accurate recall, it increases memory usage and may misrepresent data distribution. Generative models \cite{goodfellow2020generative, kingma2013auto} generate synthetic data, reducing memory usage and addressing privacy concerns while preserving task-specific features. While widely successful in computer vision \cite{shin2017continual, nguyen2017variational}, applying generative models to time-series sensor data in HAR presents challenges. Ye et al. \cite{ye2021continual} applied Deep Generative Replay (DGR) with Generative adversarial networks (GANs) to HAR tasks, but this incurs high computational costs and requires separate GANs and classifier for each task, contradicting the CL paradigm of using a singular classifier. Their approach also risks overfitting due to increased model complexity particularly with limited data. HAR-GAN showed encouraging results but faced scalability issues and data imbalance despite attempts to address them. 

Jha et. al \cite{jha2021continual} benchmarked various CL approaches on HAR datasets, highlighting LUCIR (Learning a Unified Classifier Incrementally via Rebalancing) \cite{hou2019learning} for its effectiveness in addressing data imbalance through cosine normalization. However, most studies focus on low-dimensional data with fine-grained features, lacking a systematic feature extraction process. While generative models are effective, they struggle with high-dimensional raw sensor data, which also varies due to individual differences \cite{lara2012survey} and sensor quality \cite{ye2016detecting}. Despite numerous CL applications across domains, there is limited research on applying CL strategies to raw sensor-based HAR data.


\section{Proposed Approach}
\label{sec:vae_based_framework}

\subsection{Overview}

In this work, we introduce TaskVAE, a framework for replay-based continual learning in class-incremental learning settings. In continual learning, the model begins by learning a limited set of classes (a task) and progressively incorporates new classes over time. TaskVAE uses task-specific Variational Autoencoders (VAEs) to generate exemplars from previous tasks, which are then used to train the CL classifier alongside new task data. By exposing the model to raw sensor data from new activities in each task and generating synthetic exemplars of past activities, the approach facilitates knowledge retention and enhances model adaptability to new, unseen tasks.

In a nutshell, each task is associated with its own VAE, enabling a flexible framework that can handle an increasing number of tasks without prior knowledge of the total class count. This method contrasts with approaches that require the total class count to be known upfront or rely on a single, finely-tuned VAE for all tasks. TaskVAE also introduces the flexibility to adapt to new tasks without predefining the class distribution, which makes it applicable to a wide variety of datasets, as long as tasks share the same number of classes. Furthermore, training a single generator for all tasks comes with a cost, as it faces the same CL challenges (e.g., the plasticity-stability dilemma) as the classifier. However, by training a separate VAE for each task, we mitigate this issue.

Our VAE architecture consists of an encoder, decoder, and classifier (for sample labeling). By assigning each VAE to a single task, we reduce generation complexity through an implicit divide-and-conquer approach, where each VAE is tasked with solving a conditional generation problem for a smaller, more manageable set of classes.

Our approach also incorporates a filtering mechanism that ensures only synthetic samples meeting a predefined confidence threshold are included in the training process. This combination of synthetic data generation and filtering provides a reliable solution to the challenges of knowledge retention and class-incremental learning, particularly in the context of high-dimensional sensor data in Human Activity Recognition (HAR) tasks.

Finally, a key advantage of our approach is the simplification of memory management. Unlike other CL methods that require the specification of a memory budget for storing samples, our approach has a predictable memory requirement: a single VAE per task. Since the VAE architecture is predefined and fixed, it is straightforward to quantify the memory budget required for deployment, avoiding the complexities associated with sampling-based replay approaches.

\subsection{Model architectures: VAE and CL classifier}

\begin{figure}[htbp]
\centerline{\includegraphics[scale=0.20]{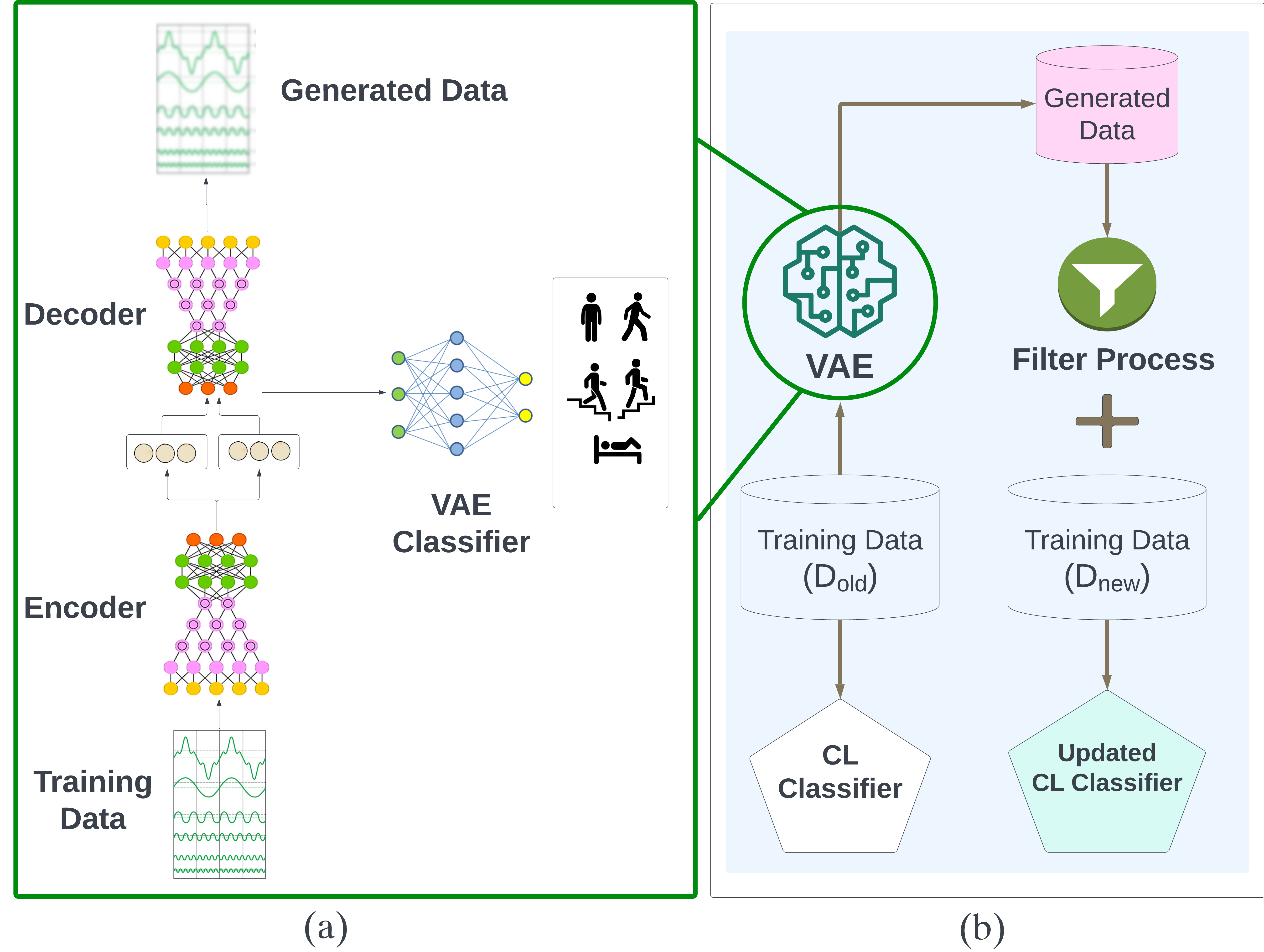}}
\caption{(a). VAE architecture (b). TaskVAE framework}
\label{fig1:vae_archi}
\end{figure}

As previously outlined, we construct a VAE with three main components: an encoder, a decoder, and a classifier as displayed in Fig. \ref{fig1:vae_archi}(a). The encoder, only used during training, aims at creating a low dimensional latent space that represents essential features of the data. Concretely the encoder produces, from each training sample, a mean and variance parameter to be fed to a normal distribution allowing a consistent exploration and organization of the latent space. The sampled latent vectors are passed to the decoder, which reconstructs the input data. The VAE classifier, built upon the latent representation, is used for labeling reconstructed data. This classifier also imposes additional structure on the latent space, enhancing feature distinction between different classes. This dual capability makes VAEs valuable for generating meaningful and diverse samples with accurate labels. During the exploitation phase, latent vectors are randomly sampled from the latent space, and the corresponding decoded vectors are used as synthetic data. These generated samples are then combined with data from the new task to train the CL classifier throughout the CL training process.

The encoder consists of five 1D convolutional layers, each followed by batch normalization and LeakyReLU activation. Max-pooling operations with the kernel size and stride of 2 are also applied between each convolutional layer to reduce the spatial dimensions of the feature maps. In the first 3 convolutional layers, a kernel size of 3 is used and the number of kernels is arranged as 16, 32, and 64 respectively. Finally, the final 2 convolutional layers have 64 filters with a kernel size of 5. The tensor is then fed into two separate fully connected layers with 64 neurons to produce the mean and log variance of the latent distribution, denoting the parameters for the probabilistic encoding of the inputs.

The decoder begins with a fully connected layer with the size of 384. After that, another series of three transposed convolutional layers, each equipped with LeakyReLU activation. The first two layers use 16 filters with the kernel sizes of 5 and 3. The final transposed convolutional layer has 6 filters and a kernel size of 3. 

Lastly, the VAE classifier is a MLP (multilayer perceptron) with a single hidden layer of 32 neurons with LeakyReLU activation. The softmax activation function is applied in the final layer to obtain a probability distribution over the class labels. 

On the other hand, the CL classifier, which is responsible for learning tasks incrementally, is a Convolutional Neural Network (CNN) with four convolutional layers with 16, 32, 32, 64 filters respectively. The first two layers use a filter size of 3, and the following two layers use a filter size of 5. Each convolutional layer is equipped with a batch normalization, ReLU activation and a max-pooling operation using a filter size of 2. Finally, the tensor is fed into the fully connected layer with 32 units, followed by the final output layer.

\subsection{Framework process: filtering mechanism in generated sample selection}

For each task in CL training, a VAE is trained individually as described in Section \ref{sec:cl_gen_data_vae_filter}. The proposed process is designed to enhance the fidelity of the generated data, ensuring that the synthetic samples more accurately represent each class. By employing a filtering mechanism, which selects only those samples with a classification confidence above a predefined threshold, $p$, from the VAE classifier, this strategy effectively reduces the inclusion of ambiguous samples. This filtering approach improves the overall quality and reliability of the pseudo-samples, thereby optimizing both the learning efficiency and predictive performance of the model. This proposed framework is illustrated in Fig. \ref{fig1:vae_archi}(b).

\subsection{CL training with generated data from TaskVAE}
\label{sec:cl_gen_data_vae_filter}
To properly assess the performance of our approach, we designed multiple CL scenarios. Each scenario consists of a series of tasks, where each task involves a subset of randomly selected classes from the unseen classes. For example, the scenario (4-5-2) consists of three tasks: the first task exposes the classifier to 4 classes, the second to 5 classes, and the third to 2 classes. A full list of the analyzed scenarios can be found in Section \ref{sec:scenarios}. For each task, a VAE is trained to generate samples specific to that task. The CL classifier is then trained on both the current task's data and the synthetic data generated by the VAEs for all preceding tasks.

In the data generation process, the strategy used for sampling latent vectors from the latent space consist on setting fixed boundaries using the minimum and maximum values for each latent space dimension, enhancing data diversity but potentially including lower-quality samples. Latent vectors are generated by sampling random uniform values within these predefined ranges.

The sampled latent vectors are input into the VAE classifier for the labeling process, and further into the decoder for generating pseudo-samples. Finally, the filtering process weeds out low quality samples.

\begin{figure}[]
\centerline{\includegraphics[scale=0.24]{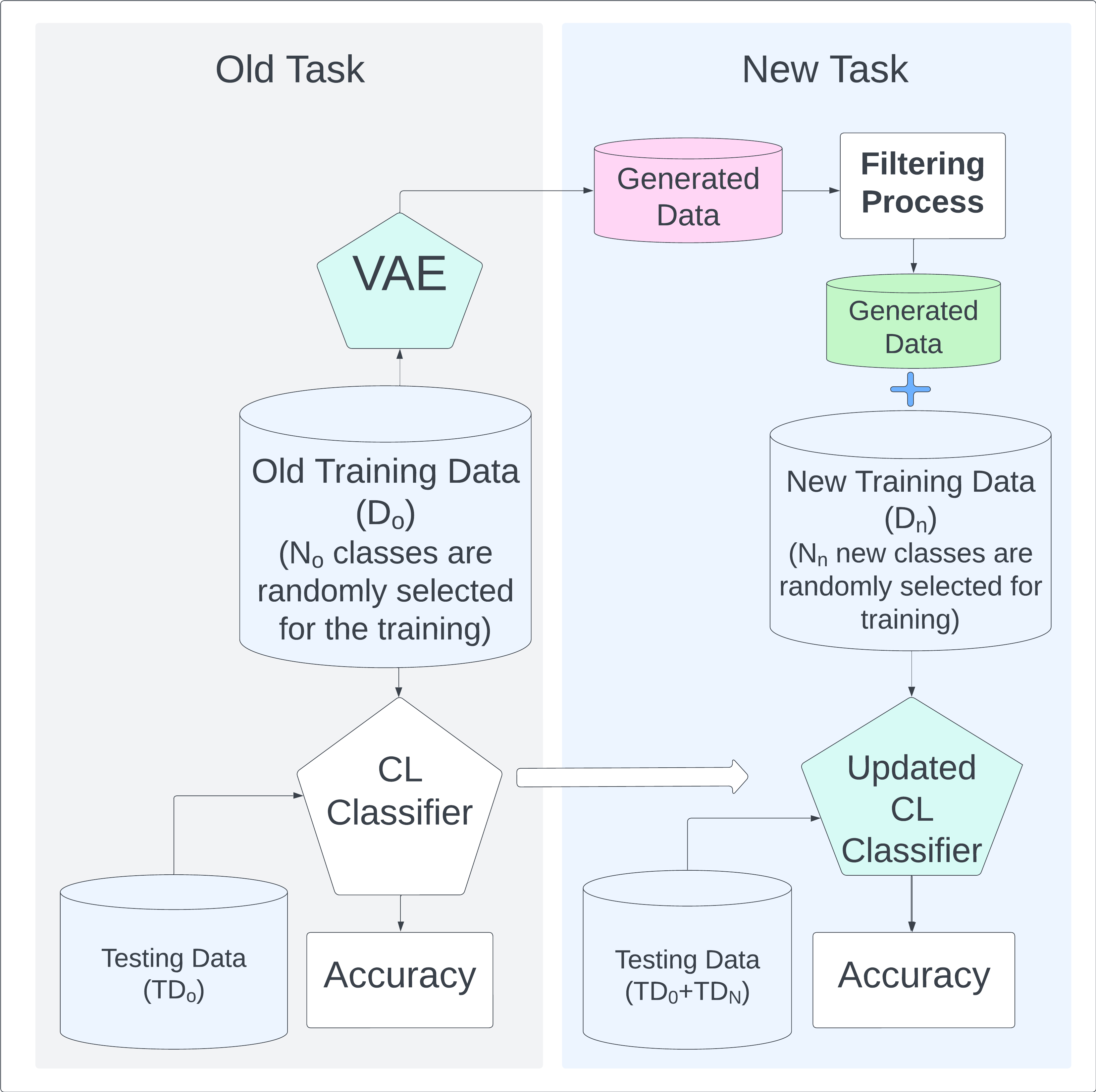}}
\caption{Training process with VAE as a generative model.}
\label{fig4:cl_vae}
\end{figure}

When the data generation process is completed, the generated data is merged with the training data of the new classes in the current task. After training, the model is evaluated using a test set that includes all the classes previously encountered during the training so far. The training process with generated data from VAE is displayed in Fig. \ref{fig4:cl_vae}.

\subsection{Evaluation strategy}
To thoroughly evaluate the effectiveness and generalizability of our approach, we conducted experiments across three distinct participants, multiple episodic learning scenarios, and various datasets. This diverse evaluation strategy ensures that our method can adapt to different individuals' activity patterns, handle incremental learning challenges (varying in classes per task but also number of tasks), and generalize across varying data distributions. 

To further validate the efficient exemplar generation performance of our method in CL training for HAR, we also compared it against state-of-the-art CL approaches sampling from real data. 
Additionally, the sampling from real data was conducted with varying holdout sizes, simulating different conditions of memory retention (often limited in CL scenarios), including memory budgets equivalent and superiors to the memory required for our method. This provides insights into how well our approach offers an alternative to memory management challenges. Finally, the performance of the model is also evaluated both with and without the filtering process.



\section{Experimental details}
\label{sec:evaluation}
This section details the materials, methods, parameters, and other relevant aspects necessary for reproducing the evaluation strategy outlined in the previous section.

\subsection{Datasets}
The study was conducted on five publicly available HAR datasets. We solely focused on accelerometer and gyroscope data from smartphones positioned at the waist, as these were the common sensor modalities and placement across all the datasets. However, they have other unique traits distinguish them from one another. The HHAR dataset \cite{stisen2015smart}, comprising sensor readings from four different smartphone models, was selected to assess our model's ability to handle data heterogeneity across different devices. The RealWorld dataset \cite{sztyler2016body}, collected from smartphones in uncontrolled environments, was chosen to test the robustness of our method under real-world conditions. The PAMAP2 dataset \cite{reiss2012introducing}, was selected for its large number of distinct activities (18), which provided a diverse and comprehensive set of movement patterns. The Motion Sense dataset \cite{Malekzadeh2019}, collected using a dedicated smartphone (iPhone 6) under a natural but controlled setting (same environment, same conditions), was included to provide a specific context for evaluating activity recognition in a controlled device environment. Finally, the UCI HAR dataset \cite{anguita2013public}, widely regarded as a standard benchmark, was chosen to provide a baseline for comparing the performance of our method on a well-established dataset with relatively simple activity patterns.

To ensure consistency across datasets, sensor data were down-sampled to 50 Hz, as recommended in \cite{sousa2019human} for HAR sensors on smartphones. Following methodologies used in previous studies \cite{IGNATOV2018915, usmanova2022federated, pagnahar2025}, data was segmented into 128-sample window ($\approx$2.56 seconds) with 50\% overlap, accelerometer and gyroscope signals. Data was then segmented by participants, with three users (with complete activity records) randomly selected. For each dataset, classes, were introduced in a class incremental learning (CIL) scenario, with 80\% of data used for training and 20\% for testing. A further 10\% of the training set was used as a validation set.

\subsection{Settings}
The experiments were conducted 30 times on a high-performance computing cluster equipped with four NVIDIA GTX 1080 Ti GPUs (11GB each) and two Intel Xeon E5-2620 v4 processors running at 2.10 GHz. Unless otherwise specified, the results presented are averages over these 30 runs. All experiments were implemented using the PyTorch framework.

\begin{table}[]
\centering
\caption{Parameter settings for the experiments}
\label{tab4:params}
\begin{tabular}{@{}ll@{}}
\toprule
\textbf{Parameters} & \textbf{Value} \\ \midrule
Learning Rate & 0.0005 \\
Batch Size & 64 \\
Number of Epochs & 20 \\
Latent Space Dimension (VAE) & 64 \\
Coefficient of Reconstruction Loss (VAE) & 1 \\
Coefficient of KL Divergence Loss (VAE) & 0.001 \\
Coefficient of Classification Loss (VAE) & 1 \\ \bottomrule
\end{tabular}
\end{table}

\subsection{Learning scenarios}
\label{sec:scenarios}
Experimenting with multiple episodic class incremental learning scenarios in CL offers a realistic assessment of how models perform in dynamic, real-world environments, where new data arrives sequentially and the system must adapt without retraining from scratch.  We aimed to explore a broad spectrum of scenarios. However, the sheer number of potential scenarios was too vast to consider them all. As a result, we made an arbitrary selection of representative scenarios, prioritizing more difficult scenarios and ensuring that we covered a wide variety of situations while still maintaining a manageable scope for experimentation. The scenarios are summarized in Table \ref{tab:scenarios}. For the UCI HAR 5-task scenario, the data was insufficient, making it irrelevant to conduct experiments.

As a note, it is important to highlight that the earlier in the scenario tasks with a large number of classes are introduced, the more challenging it becomes for the model to adapt and maintain performance.

\setlength{\tabcolsep}{1.5pt} 
\def\arraystretch{0.75}
\begin{table}[]
\centering
\caption{Scenario settings (Dashed boxes represent impossible scenarios due to class number limitations)}
\label{tab:scenarios}
\begin{tabular}{@{}cccccc@{}}
\toprule
\multicolumn{1}{l}{} & \multicolumn{5}{c}{\textbf{Dataset}} \\ \midrule
\multicolumn{1}{l}{} & \textbf{PAMAP2} & \textbf{HHAR} & \textbf{RealWorld} & \textbf{MotionSense} & \textbf{UCI HAR} \\
 &  &  &  &  &  \\
 &  &  &  &  &  \\
 & \multirow{-3}{*}{(6-4)} & \multirow{-3}{*}{\begin{tabular}[c]{@{}c@{}}(3-3)\\ (4-2)\end{tabular}} & \multirow{-3}{*}{(5-3)} & \multirow{-3}{*}{\begin{tabular}[c]{@{}c@{}}(3-3)\\ (4-2)\end{tabular}} & \multirow{-3}{*}{\begin{tabular}[c]{@{}c@{}}(3-3)\\ (4-2)\end{tabular}} \\
 &  &  & &  &  \\
 &  &  & &  &  \\
 & \multirow{-3}{*}{\begin{tabular}[c]{@{}c@{}}(4-3-3)\\ (5-3-2)\end{tabular}} & \multirow{-3}{*}{\begin{tabular}[c]{@{}c@{}}(2-2-2)\\ (2-3-1)\\ (3-2-1)\end{tabular}} & \multirow{-3}{*}{\begin{tabular}[c]{@{}c@{}}(3-3-2)\\ (4-2-2)\end{tabular}} & \multirow{-3}{*}{\begin{tabular}[c]{@{}c@{}}(2-2-2)\\ (2-3-1)\\ (3-2-1)\end{tabular}} & \multirow{-3}{*}{\begin{tabular}[c]{@{}c@{}}(2-2-2)\\ (2-3-1)\\ (3-2-1)\end{tabular}} \\
 &  &  &  &  &  \\
 &  &  &  &  &  \\
 & \multirow{-3}{*}{\begin{tabular}[c]{@{}c@{}}(4-3-2-1)\\ (2-3-4-1)\end{tabular}} & \multirow{-3}{*}{(3-1-1-1)} & \multirow{-3}{*}{\begin{tabular}[c]{@{}c@{}}(2-2-2-2)\\ (4-2-1-1)\end{tabular}} & \multirow{-3}{*}{(3-1-1-1)} & \multirow{-3}{*}{(3-1-1-1)} \\
 &  &  &  &  & \\
 &  &  &  &  & \\
 & \multirow{-3}{*}{(2-2-2-2-2)} & \multirow{-3}{*}{(2-1-1-1-1)} & \multirow{-3}{*}{\begin{tabular}[c]{@{}c@{}}(4-1-1-1-1)\\ (2-3-1-1-1)\\ (3-2-1-1-1)\end{tabular}} & \multirow{-3}{*}{(2-1-1-1-1)} & \multirow{-3}{*}{¤ ¤} \\
 &  &  &  &  &  \\
 &  &  &  &  & \\
\multirow{-15}{*}{\rotatebox[origin=c]{90}{\textbf{Scenarios}}} & (5-1-1-1-1-1) & --- & --- & --- & ---
\end{tabular}%
\end{table}

\subsection{Comparison methods}
\begin{figure}[]
\centerline{\includegraphics[scale=0.26]{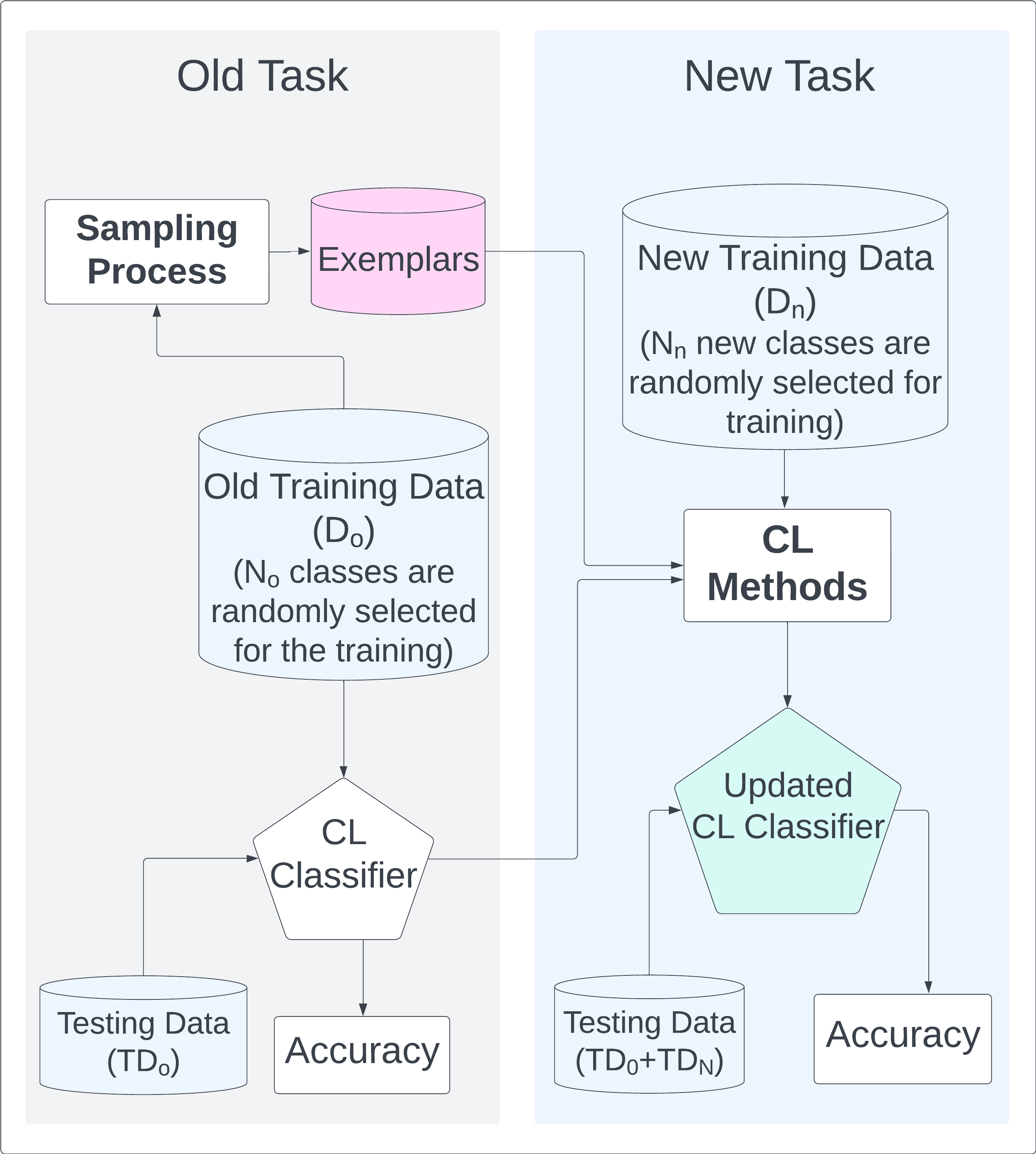}}
\caption{Training process with CL methods using real data as exemplars}
\label{fig3:cl_real_data}
\end{figure}

The training process for these methods is shown in Fig. \ref{fig3:cl_real_data}.

\paragraph*{Naive baseline method - CL training with random sampling from the real data}
At the end of each task, it chooses $N$ exemplars at random from the task data where $N = S/T$ ($S$ the exemplar size and $T$ the number of tasks) and saves them to memory.
\paragraph*{Hybrid CL Methods using real data} The model from the previous task is also used for different processes in each CL approach. The CL approaches selected for these experiments are Elastic Weight Consolidation with Replay (EWC Replay) \cite{kirkpatrick2017overcoming} that uses a random sampling, Incremental Classifier and Representation Learning (iCaRL)\cite{rebuffi2017icarl}, and LUCIR\cite{hou2019learning} that use a hearding sampling\cite{welling2009herding}.


\subsection{Exemplar sizes}
We selected five different holdout sizes to explore the impact of memory allocation during sampling from real data. The first holdout size matches the VAE's memory footprint (392 KB), ensuring a direct comparison between the memory required for our approach and the memory required for the other methods. This corresponds to an exemplar size (eq. VAE) of 60/task for each dataset.

The other four sizes follow a geometric progression in four steps, ranging from 100 to the maximum number of available exemplars, which corresponds to the total size of the dataset (effectively mimicking an infinite memory budget since all samples can be committed to memory). The exemplar size is given as a total budget for the entire continual learning process. 



When generating data from VAE, we align the generated data size per class to the average training data size for each new task, trying to achieve balanced training sets. For the filtering process, previous experiments were conducted using values of $p$ within the range [0.75, 0.97], with the best performance typically observed at $p = 0.8$ for some datasets. However, for more complex datasets, we were not always able to obtain samples at this value. To ensure comparability across all datasets, a threshold value of $p = 0.60$ was selected, reflecting the goal of excluding noisy or outlier data.


\subsection{Metrics}
In CL context, the model is trained with a continually updating data stream of multiple tasks. Hence, it is necessary to conduct a comprehensive evaluation which not only assesses the model's capability to adapt to new information but also to retain past knowledge. In this experiment, we focus on the following metrics:
\paragraph*{Accuracy by Tasks (ACT)}
It is the accuracy of recognizing all classes trained so far in each task. This metric is used to indicate the overall performance of the model with both newly and previously learned classes in all tasks.
\paragraph*{New-Class Accuracy by Tasks (NCT)} It is the accuracy of recognizing new activities in the current task. This metric helps evaluating plasticity: the model's ability to learn new information.  
\paragraph*{Old-Class Accuracy by Tasks (OCT)} It is the accuracy of recognizing all old classes which have been learnt from the previous tasks. It is a measure of the model's stability: the model's ability to retain knowledge of previously learned classes when trained on new tasks.

\section{Results \& Discussion}
\label{sec:results}
Due to space limitations, only representative results are presented, capturing overall trends for discussion. The complete results and source code are available at \url{https://github.com/bonpagnakann/TaskVAE}.

Table \ref{tab:summaryresultsP0} summarizes the results for one participant in all scenarios, all datasets, all holdout sizes, all methods and TaskVAE with and without filter. These results trends are confirmed for participants P1 and P2 (see repository link). It is evident that TaskVAE achieves the best overall performance across the scenarios performing better in 11 out of the 35 instances. In contrast, TaskVAE without the filter performs better on 6 instances. There are 15 instances where both versions show equal performance and outperform all other approaches. This highlights the consistent advantage of the filter, but also suggests that, in certain cases, the filter may not make a significant difference.

\begin{figure*}[h!]
    \centering
    \begin{subfigure}[b]{0.32\textwidth}
        \centering
        \includegraphics[width=\linewidth]{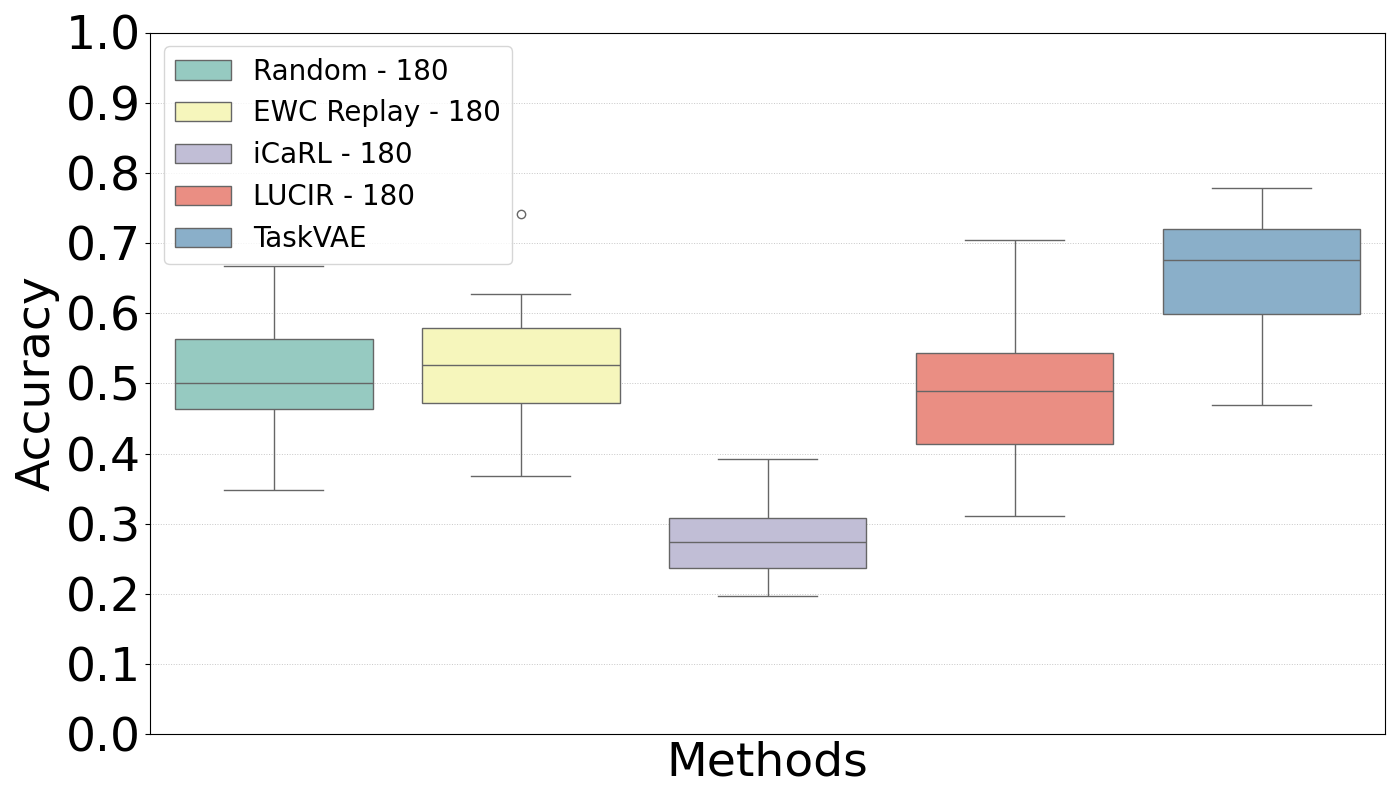}
        \caption{UCI HAR}
    \end{subfigure}
    \begin{subfigure}[b]{0.32\textwidth}
        \centering
        \includegraphics[width=\linewidth]{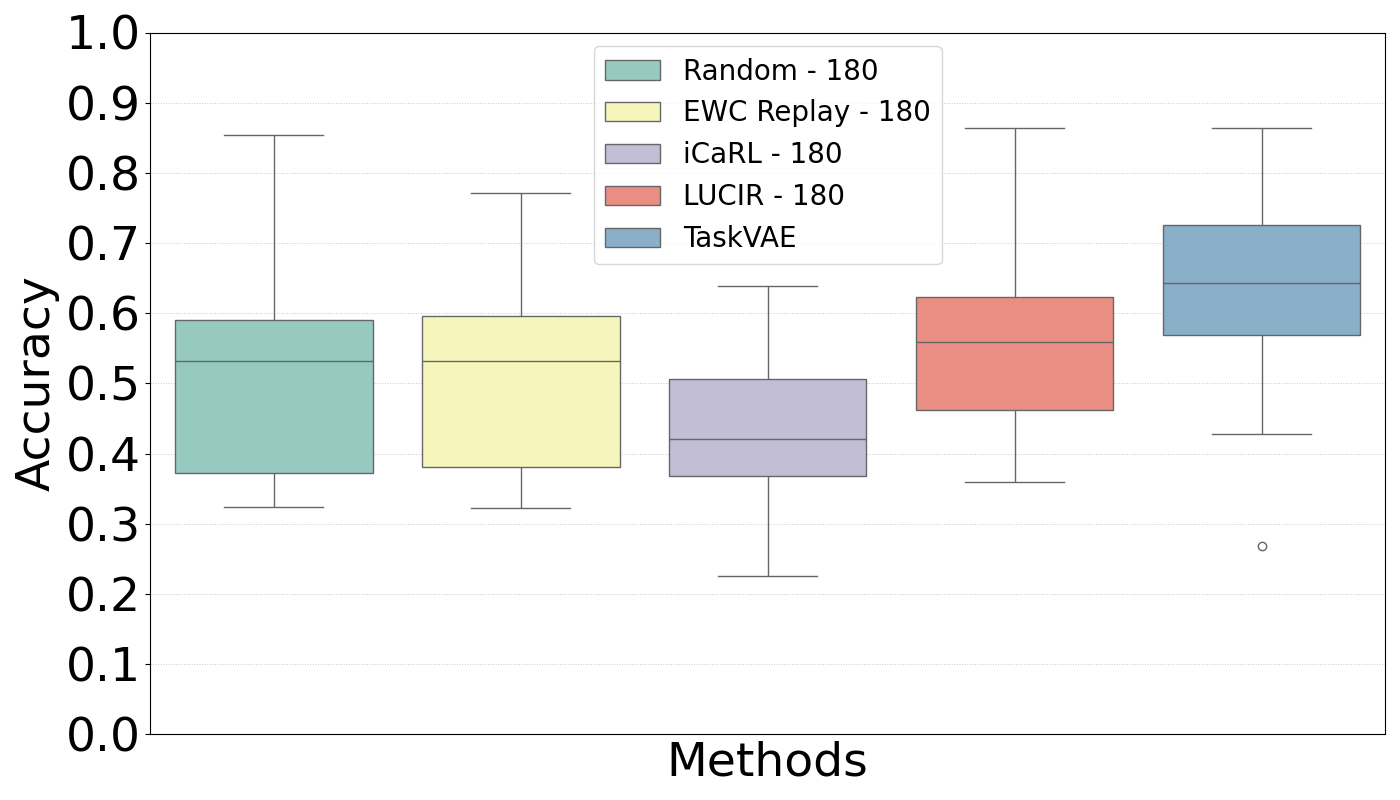}
        \caption{MotionSense}
    \end{subfigure}
    \begin{subfigure}[b]{0.32\textwidth}
        \centering
        \includegraphics[width=\linewidth]{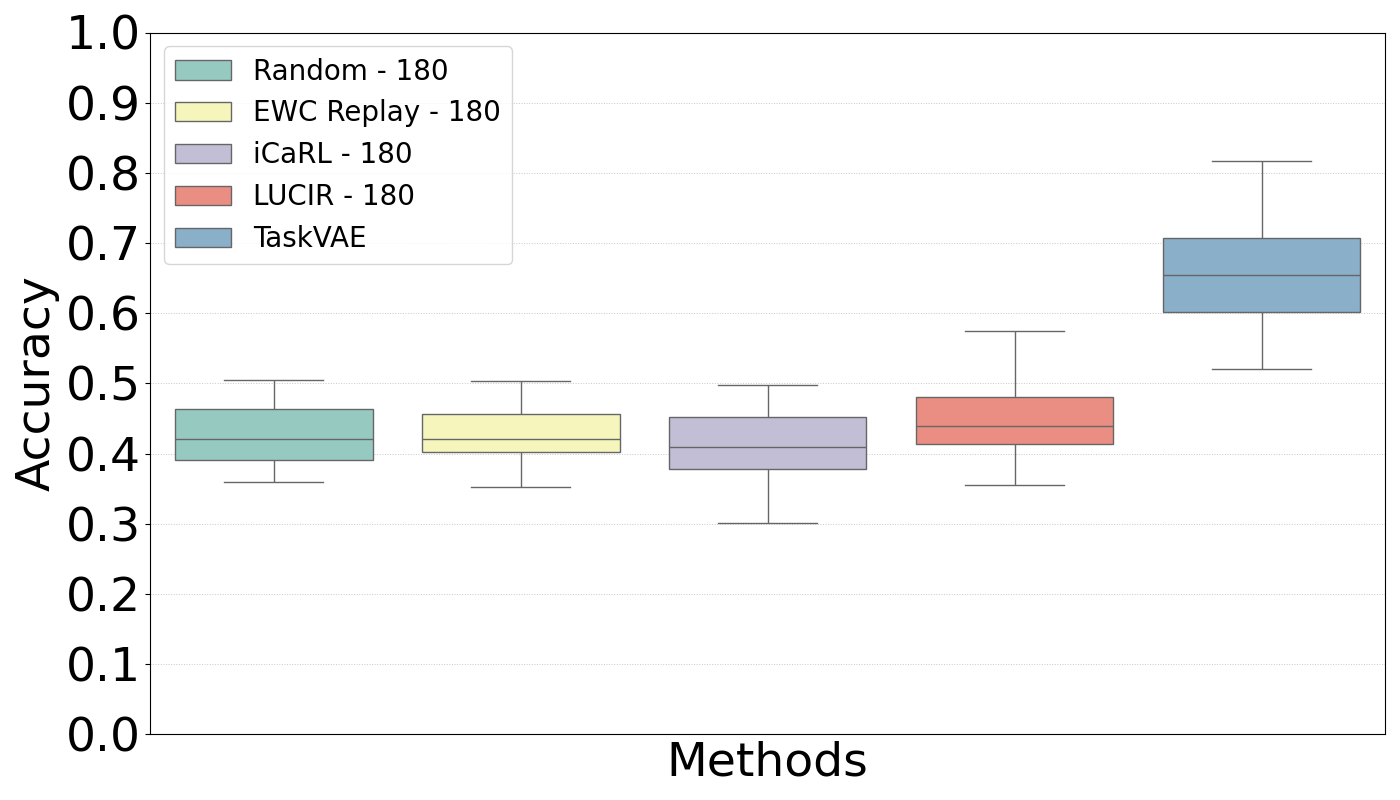}
        \caption{HHAR}
    \end{subfigure}
    \caption{Accuracy box plots by tasks (30 runs) on the last task for participant P0 in a 3-task scenario (2-2-2), exemplar size eq. VAE (60/task), methods (left to right): Random, EWC-Replay, iCaRL, LUCIR, TaskVAE.}
    \label{fig:boxplot_all_acc}
\end{figure*}

Fig. \ref{fig:boxplot_all_acc} compares accuracy box plots by task on the last task of participant P0 in a 3-task scenario (2-2-2), with an exemplar size set to 60 per taks (eq. VAE). Each box plot represents the accuracy variability across 30 runs for each method (namely, Random, EWC-Replay, iCaRL, LUCIR and TaskVAE). From this figure, we can observe that TaskVAE exhibits the least variability indicating that its performance is stable across different runs. The other methods show wider boxes, highlighting a greater degree of variability, this is particularly pronounced in the random approach.

\begin{figure*}[h!]
    \centering
    \begin{subfigure}{0.32\textwidth}
        \centering
        \includegraphics[width=\linewidth]{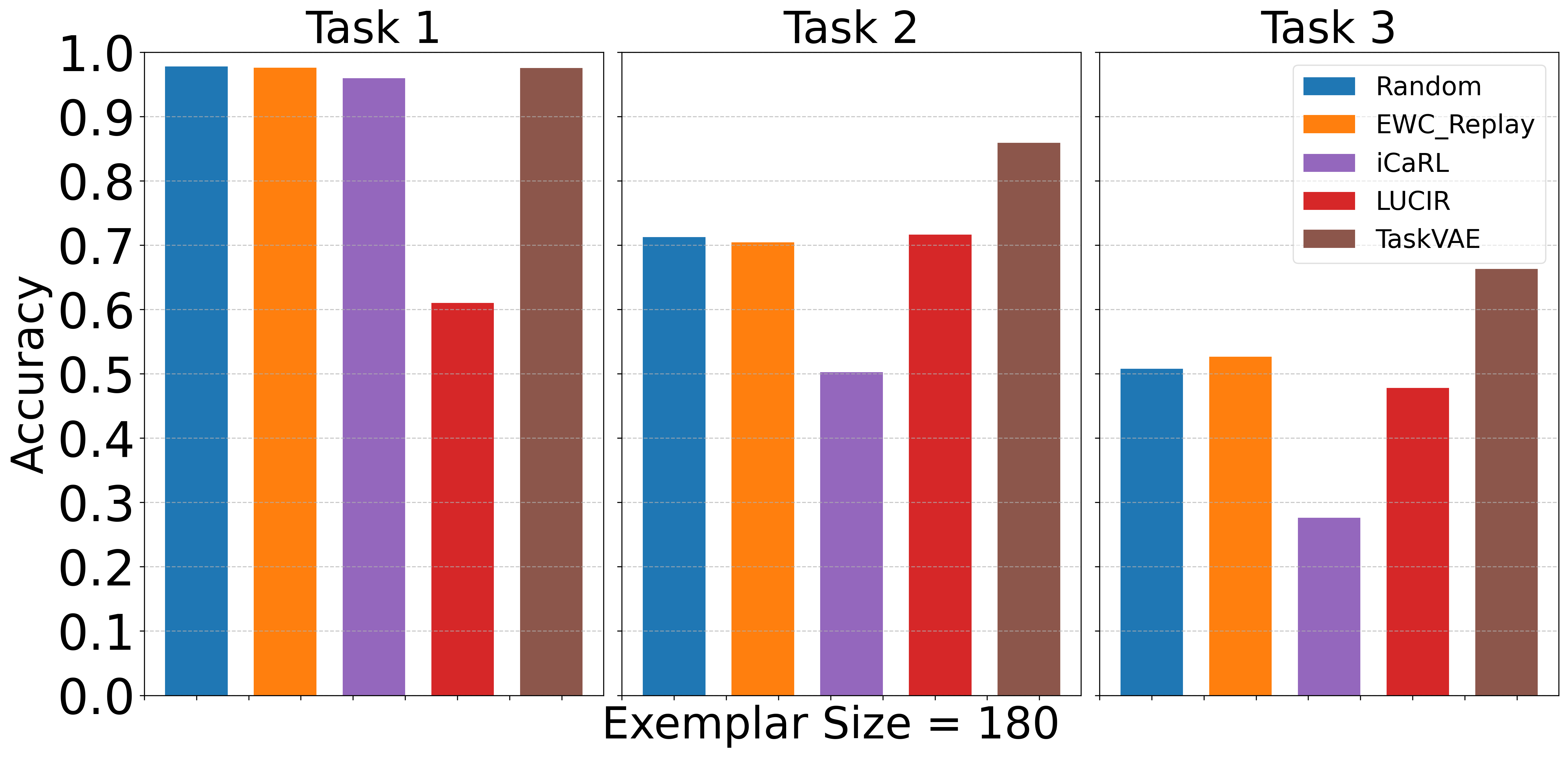}
    \end{subfigure} 
    \begin{subfigure}{0.32\textwidth}
        \centering
        \includegraphics[width=\linewidth]{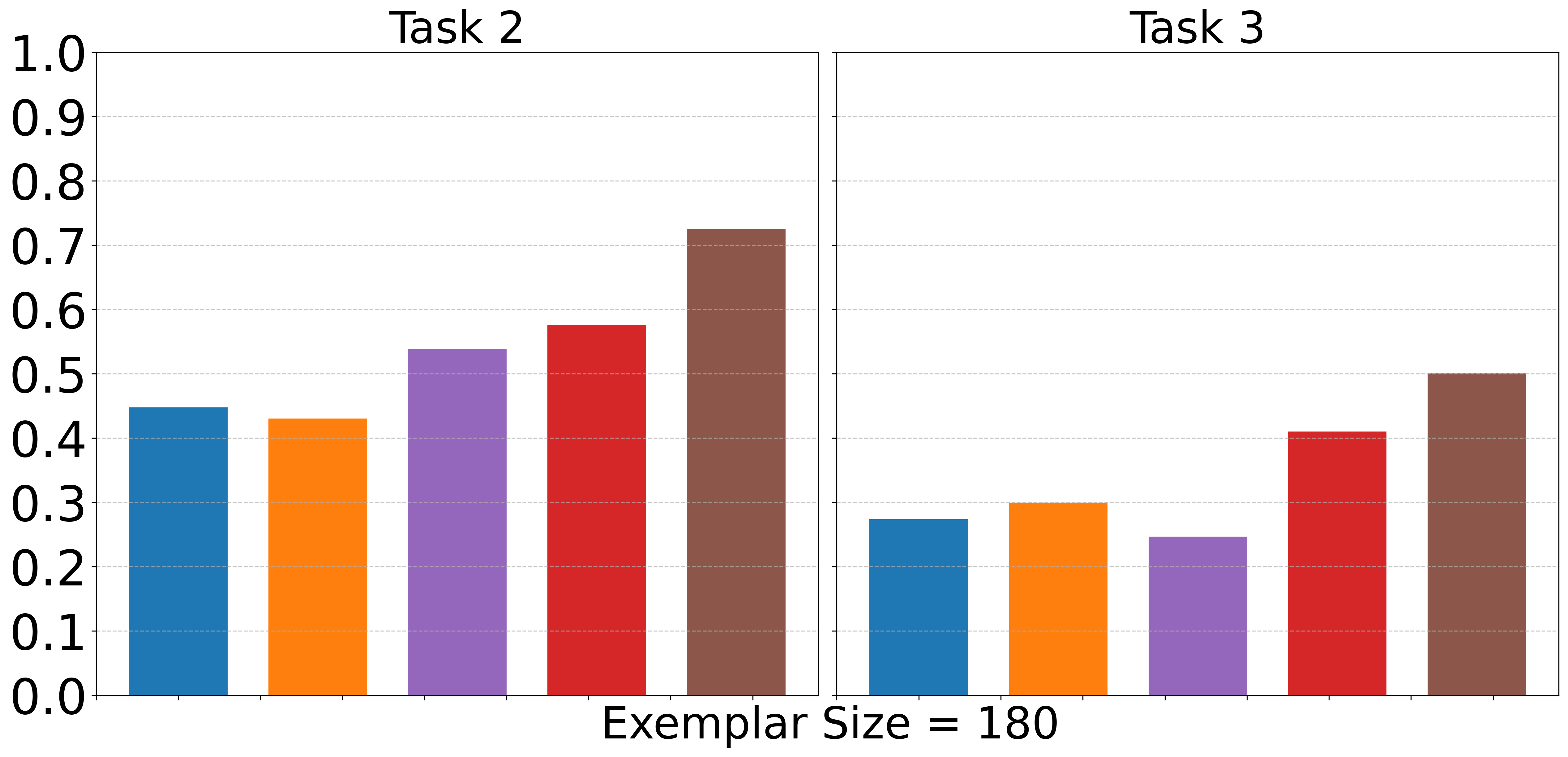}
    \end{subfigure} 
    \begin{subfigure}{0.32\textwidth}
        \centering
        \includegraphics[width=\linewidth]{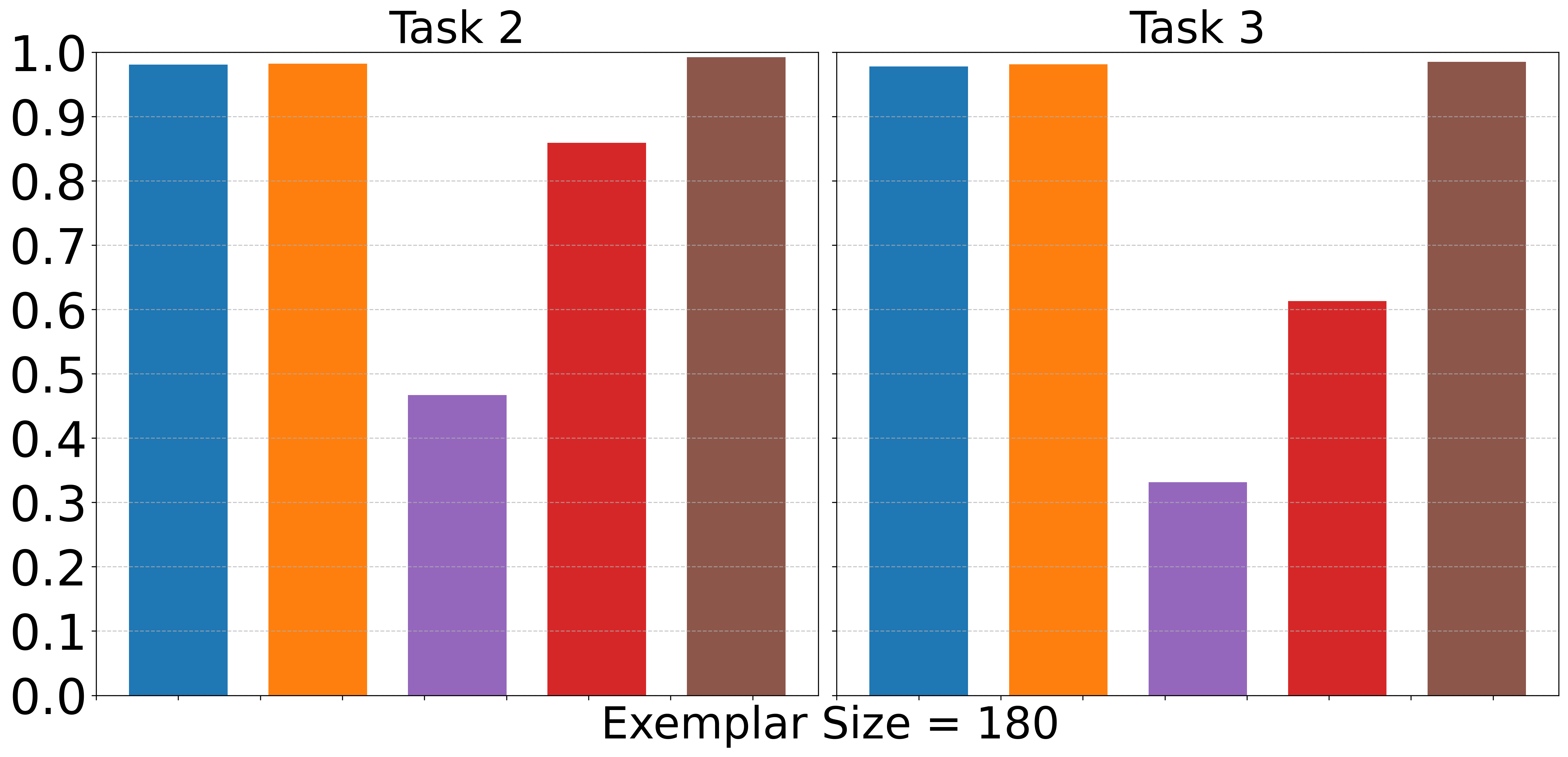}
    \end{subfigure} 

    \vspace{0.1cm} 

    \begin{subfigure}{0.32\textwidth}
        \centering
        \includegraphics[width=\linewidth]{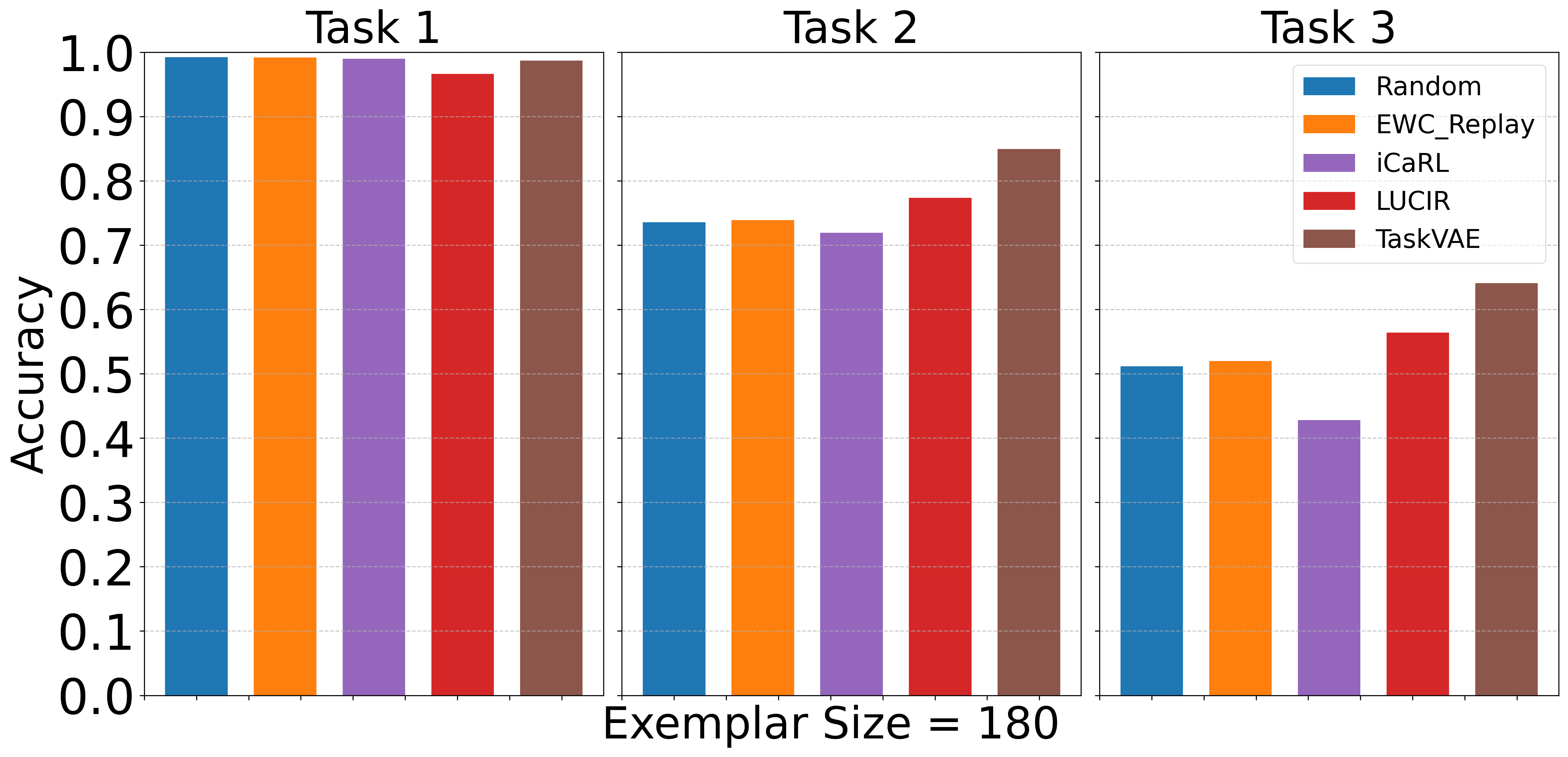}
    \end{subfigure} 
    \begin{subfigure}{0.32\textwidth}
        \centering
        \includegraphics[width=\linewidth]{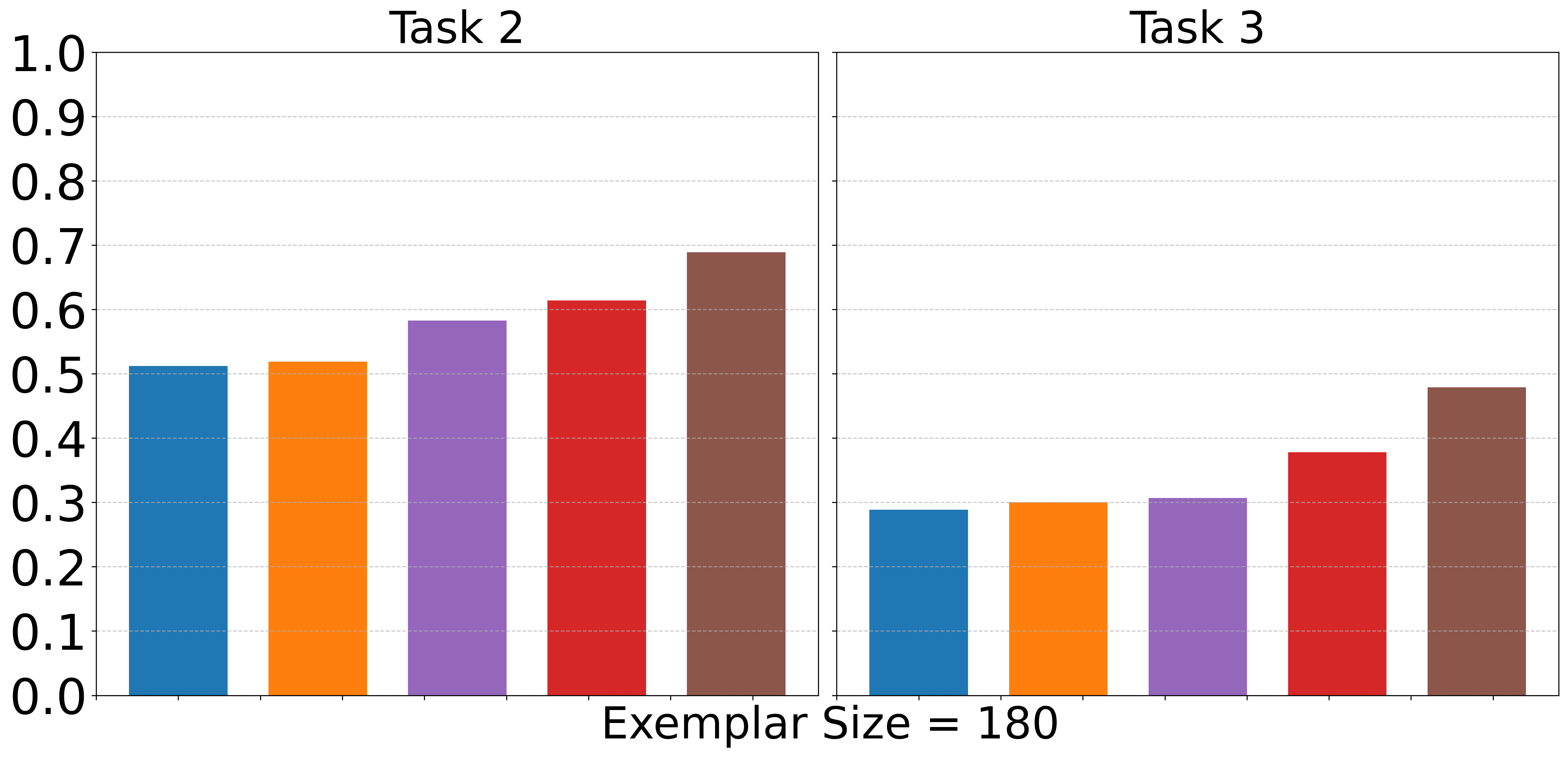}
    \end{subfigure} 
    \begin{subfigure}{0.32\textwidth}
        \centering
        \includegraphics[width=\linewidth]{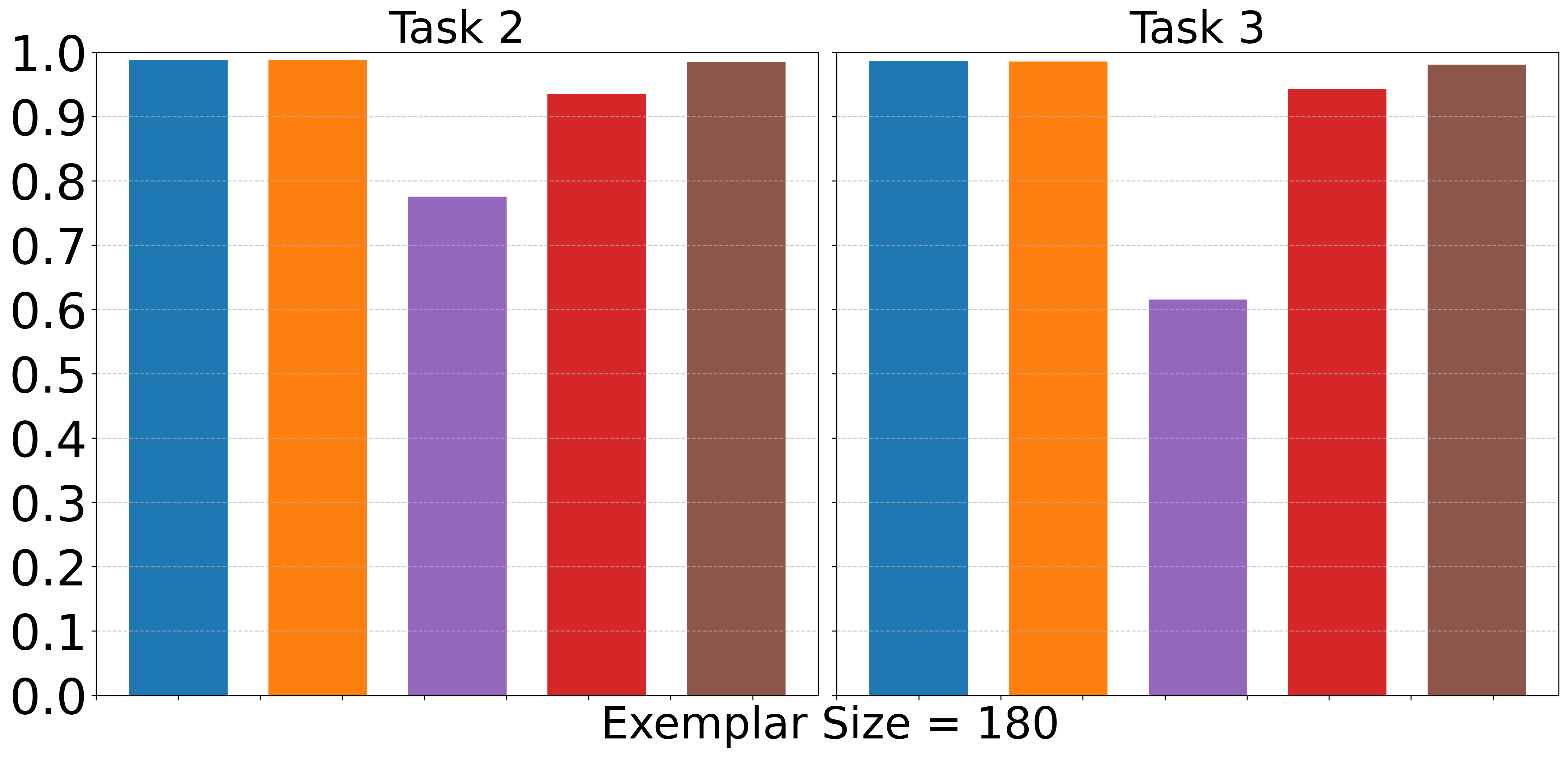}
    \end{subfigure} 

    \vspace{0.1cm} 

    \begin{subfigure}{0.32\textwidth}
        \centering
        \includegraphics[width=\linewidth]{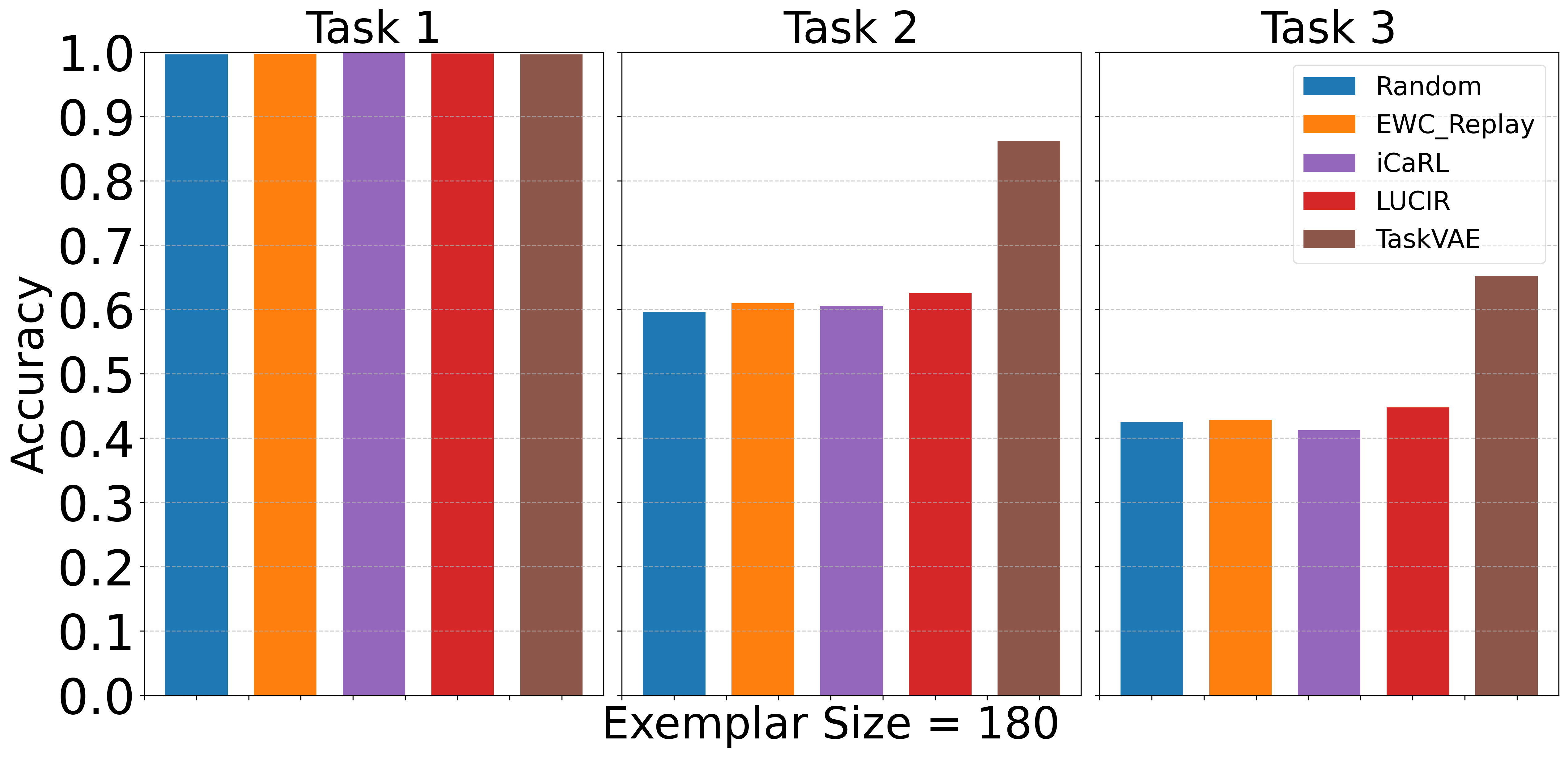}
        \caption{Accuracy by Task}
        \label{fig:all_acc}
    \end{subfigure} 
    \begin{subfigure}{0.32\textwidth}
        \centering
        \includegraphics[width=\linewidth]{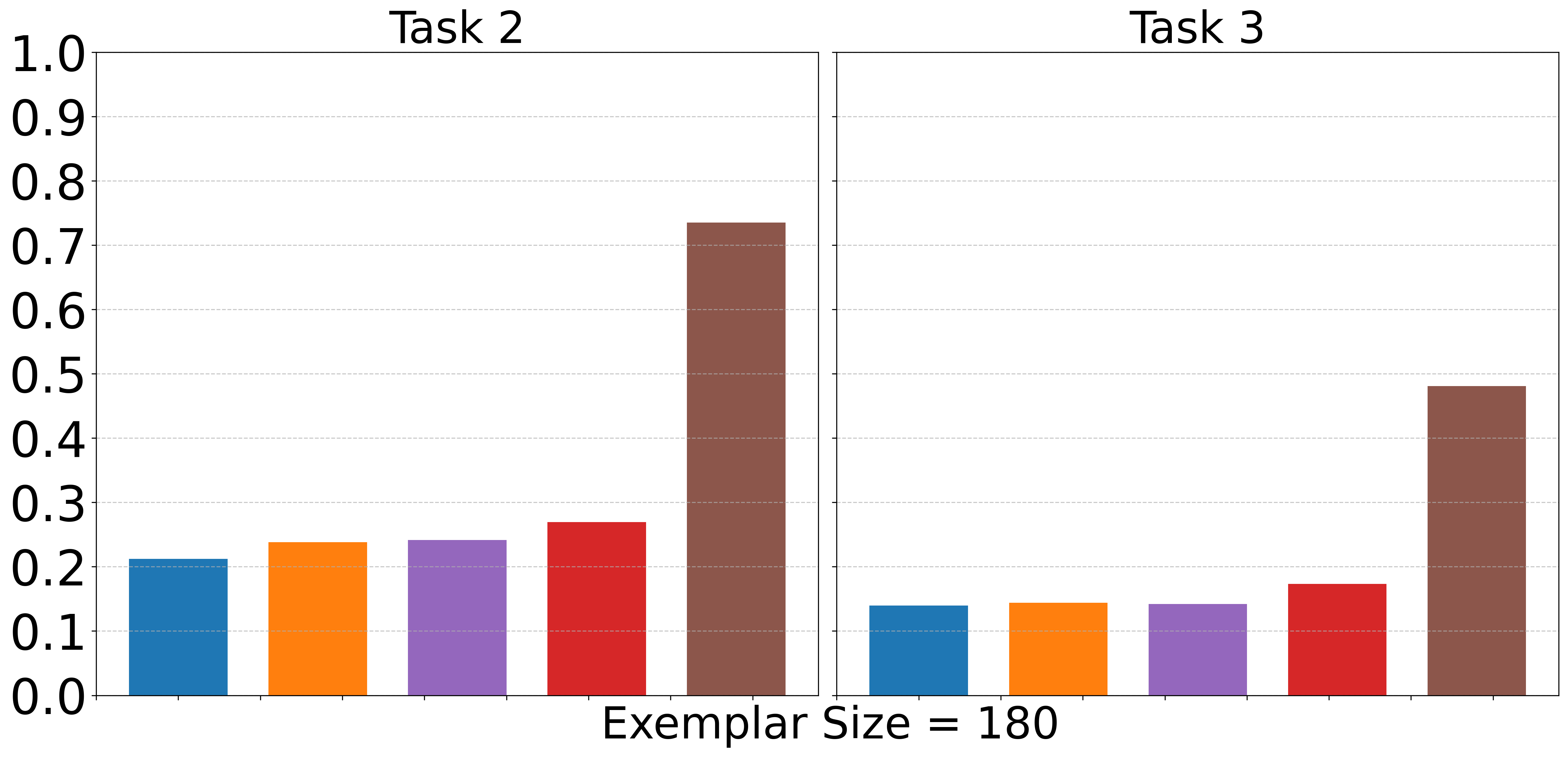}
        \caption{Old-Class Accuracy by Tasks}
        \label{fig:old_acc}
    \end{subfigure} 
    \begin{subfigure}{0.32\textwidth}
        \centering
        \includegraphics[width=\linewidth]{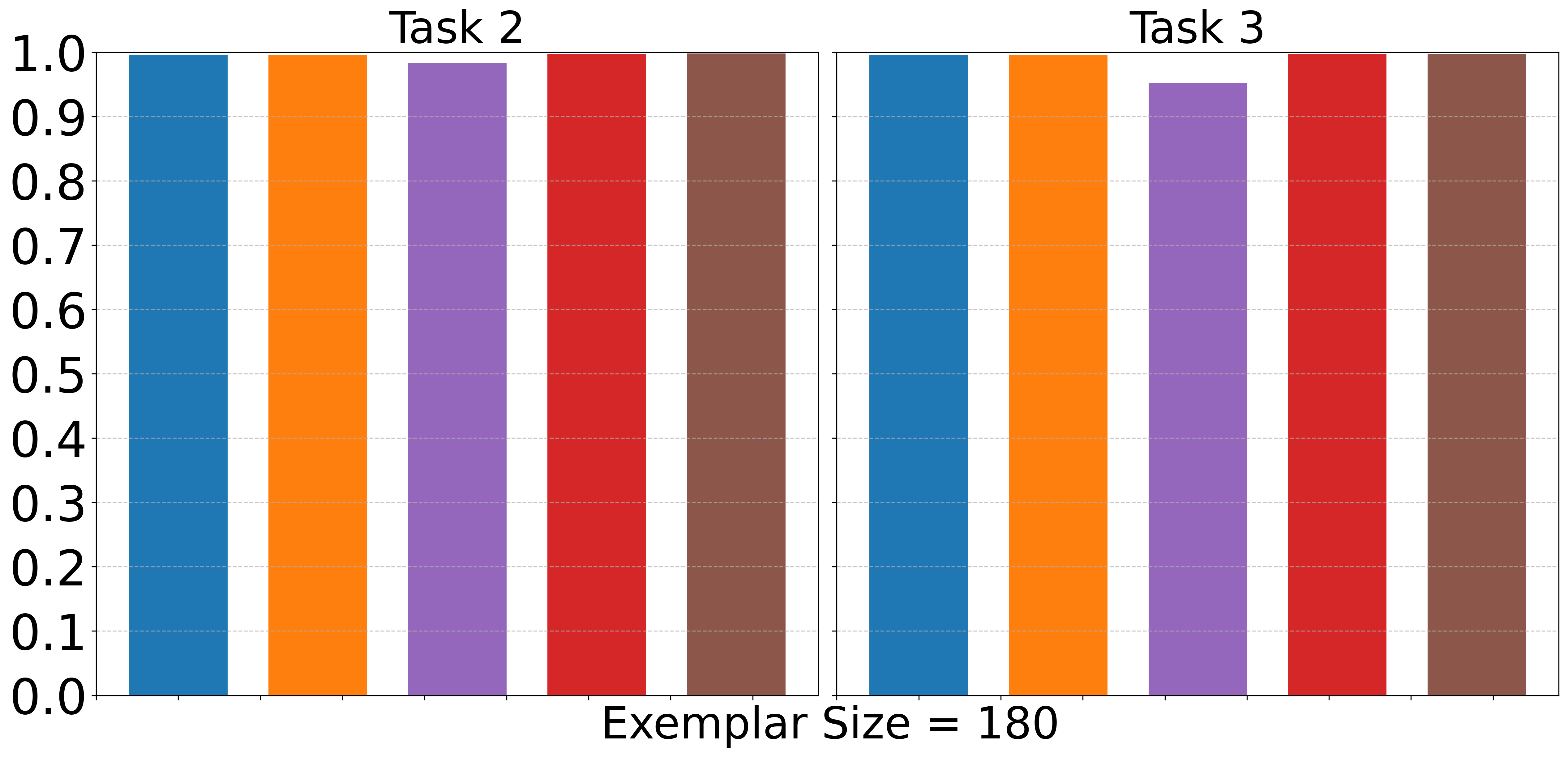}
        \caption{New-Class Accuracy by Tasks}
        \label{fig:new_acc}
    \end{subfigure}
    \caption{Average metrics by tasks (30 runs) for participant P0 in scenario (2-2-2), exemplar size eq. VAE (60/task), datasets: UCI HAR, Motion Sense, HHAR (top to bottom). Methods order per task: Random, EWC-Replay, iCaRL, LUCIR, TaskVAE (left to right).}
    \label{fig:all_metrics}
\end{figure*}

Fig. \ref{fig:all_acc} illustrates the all-class accuracy across tasks. The result show that random sampling, EWC Replay, iCaRL, and LUCIR have comparable performance, while TaskVAE outperforms the other methods. Specifically, TaskVAE maintains an accuracy around 85\% in Task 2 and approximately 65\% in Task 3. The results for TaskVAE without the filter, which follow the same trends discussed earlier in relation to Table \ref{tab:summaryresultsP0}, are not included in the figure due to space constraints.

\setlength{\tabcolsep}{1.5pt} 
\begin{table*}[]
\centering
\caption{Summary of results for P0, all scenarios, all datasets, all holdout sizes, all methods and TaskVAE with and without filter. Best performances per row in bold.}
\label{tab:summaryresultsP0}
\resizebox{\textwidth}{!}{%
\begin{tabular}{@{}|ccl|cccc|cccc|cccc|cccc|cccc|c|c|@{}}
\multicolumn{3}{c}{\cellcolor[HTML]{E8E8E8}\textbf{P0}} & \textbf{Rand} & \textbf{EWC-R} & \textbf{iCaRL} & \textbf{LUCIR} & \textbf{Rand} & \textbf{EWC-R} & \textbf{iCaRL} & \textbf{LUCIR} & \textbf{Rand} & \textbf{EWC-R} & \textbf{iCaRL} & \textbf{LUCIR} & \textbf{Rand} & \textbf{EWC-R} & \textbf{iCaRL} & \textbf{LUCIR} & \textbf{Rand} & \textbf{EWC-R} & \textbf{iCaRL} & \textbf{LUCIR} &  &  \\
 & \multicolumn{2}{c}{\cellcolor[HTML]{ADADAD}\textbf{Exemplar Size}} & \multicolumn{4}{c|}{\textbf{100}} & \multicolumn{4}{c|}{\textbf{142}} & \multicolumn{4}{c|}{\textbf{201}} & \multicolumn{4}{c|}{\textbf{286}} & \multicolumn{4}{c|}{\textbf{60/task (eq. VAE)}} & \multirow{-2}{1.5cm}{\centering\vspace{-1ex}\textbf{TaskVAE w/o filter}} & \multirow{-2}{1.5cm}{\centering\textbf{TaskVAE}} \\
\multicolumn{1}{|c|}{} & \multicolumn{1}{c|}{} & \textbf{3-1-1-1} & 0.35 & 0.35 & 0.25 & 0.33 & 0.33 & 0.33 & 0.23 & 0.35 & 0.33 & 0.33 & 0.22 & 0.34 & 0.33 & 0.35 & 0.22 & 0.34 & 0.34 & 0.33 & 0.21 & 0.33 & 0.45 & \textbf{0.47} \\
\multicolumn{1}{|c|}{} & \multicolumn{1}{c|}{} & \textbf{2-2-2} & 0.50 & 0.49 & 0.30 & 0.45 & 0.51 & 0.50 & 0.30 & 0.49 & 0.52 & 0.51 & 0.28 & 0.48 & 0.53 & 0.52 & 0.28 & 0.48 & 0.51 & 0.53 & 0.28 & 0.48 & 0.65 & \textbf{0.66} \\
\multicolumn{1}{|c|}{} & \multicolumn{1}{c|}{} & \textbf{2-3-1} & 0.42 & 0.38 & 0.28 & 0.39 & 0.40 & 0.40 & 0.26 & 0.40 & 0.41 & 0.40 & 0.25 & 0.42 & 0.42 & 0.41 & 0.26 & 0.41 & 0.40 & 0.41 & 0.27 & 0.42 & 0.65 & \textbf{0.66} \\
\multicolumn{1}{|c|}{} & \multicolumn{1}{c|}{} & \textbf{3-2-1} & 0.39 & 0.39 & 0.26 & 0.36 & 0.36 & 0.38 & 0.25 & 0.40 & 0.37 & 0.38 & 0.24 & 0.41 & 0.39 & 0.39 & 0.23 & 0.40 & 0.38 & 0.37 & 0.24 & 0.41 & \textbf{0.63} & 0.61 \\
\multicolumn{1}{|c|}{} & \multicolumn{1}{c|}{} &  \textbf{3-3} & 0.69 & 0.68 & 0.33 & 0.69 & 0.70 & 0.69 & 0.32 & 0.66 & 0.70 & 0.70 & 0.33 & 0.70 & 0.70 & 0.69 & 0.32 & 0.68 & 0.66 & 0.67 & 0.32 & 0.62 & \textbf{0.83} & \textbf{0.83} \\
\multicolumn{1}{|c|}{\multirow{-6}{*}{\rotatebox[origin=c]{90}{\textbf{UCI HAR}}}} & \multicolumn{1}{c|}{\multirow{-6}{*}{\rotatebox[origin=c]{90}{\textbf{Scenario}}}} & \textbf{4-2} & 0.55 & 0.56 & 0.34 & 0.56 & 0.57 & 0.57 & 0.32 & 0.60 & 0.58 & 0.59 & 0.34 & 0.58 & 0.59 & 0.58 & 0.32 & 0.61 & 0.57 & 0.57 & 0.35 & 0.58  & 0.77 & \textbf{0.78} \\
 & \multicolumn{2}{c}{\cellcolor[HTML]{ADADAD}\textbf{Exemplar Size}} & \multicolumn{4}{c}{\textbf{100}} & \multicolumn{4}{c}{\textbf{197}} & \multicolumn{4}{c}{\textbf{390}} & \multicolumn{4}{c}{\textbf{770}} & \multicolumn{4}{c}{\textbf{60/task (eq. VAE)}} & \cellcolor[HTML]{ADADAD}\textbf{} & \cellcolor[HTML]{ADADAD}\textbf{} \\
\multicolumn{1}{|c|}{} & \multicolumn{1}{c|}{} & \textbf{2-1-1-1-1} & 0.32 & 0.31 & 0.30 & 0.32 & 0.31 & 0.31 & 0.30 & 0.32 & 0.30 & 0.31 & 0.30 & 0.32 & 0.31 & 0.31 & 0.27 & 0.31 & 0.31 & 0.31 & 0.31 & 0.33 & \textbf{0.40} & 0.39 \\
\multicolumn{1}{|c|}{} & \multicolumn{1}{c|}{} & \textbf{3-1-1-1} & 0.32 & 0.32 & 0.32 & 0.33 & 0.31 & 0.31 & 0.31 & 0.33 & 0.31 & 0.31 & 0.32 & 0.32 & 0.32 & 0.32 & 0.28 & 0.32 & 0.31 & 0.31 & 0.31 & 0.32 & 0.44 & \textbf{0.45} \\
\multicolumn{1}{|c|}{} & \multicolumn{1}{c|}{} & \textbf{2-2-2} & 0.49 & 0.48 & 0.42 & 0.54 & 0.52 & 0.52 & 0.44 & 0.54 & 0.53 & 0.53 & 0.38 & 0.57 & 0.55 & 0.55 & 0.36 & 0.58 & 0.51 & 0.52 & 0.43 & 0.56 & \textbf{0.66} & 0.64 \\
\multicolumn{1}{|c|}{} & \multicolumn{1}{c|}{} & \textbf{2-3-1} & 0.45 & 0.45 & 0.39 & 0.49 & 0.45 & 0.45 & 0.35 & 0.49 & 0.45 & 0.45 & 0.35 & 0.47 & 0.47 & 0.47 & 0.31 & 0.50 & 0.45 & 0.44 & 0.36 & 0.48 & 0.62 & \textbf{0.65} \\
\multicolumn{1}{|c|}{} & \multicolumn{1}{c|}{} & \textbf{3-2-1} & 0.37 & 0.37 & 0.34 & 0.42 & 0.38 & 0.39 & 0.32 & 0.41 & 0.39 & 0.37 & 0.32 & 0.40 & 0.39 & 0.39 & 0.29 & 0.43 & 0.38 & 0.38 & 0.32 & 0.41 & \textbf{0.61} & \textbf{0.61} \\
\multicolumn{1}{|c|}{} & \multicolumn{1}{c|}{} & \textbf{3-3} & 0.67 & 0.67 & 0.52 & 0.76 & 0.71 & 0.71 & 0.46 & 0.76 & 0.73 & 0.73 & 0.41 & 0.78 & 0.75 & 0.73 & 0.39 & 0.79 & 0.70 & 0.70 & 0.51 & 0.75 & \textbf{0.80} & \textbf{0.80} \\
\multicolumn{1}{|c|}{\multirow{-7}{*}{\rotatebox[origin=c]{90}{\textbf{Motion Sense}}}} & \multicolumn{1}{c|}{\multirow{-7}{*}{\rotatebox[origin=c]{90}{\textbf{Scenario}}}} & \textbf{4-2} & 0.60 & 0.60 & 0.54 & 0.70 & 0.63 & 0.63 & 0.50 & 0.72 & 0.67 & 0.65 & 0.44 & 0.75 & 0.68 & 0.69 & 0.40 & 0.76 & 0.60 & 0.60 & 0.52 & 0.72 & \textbf{0.76} & 0.75 \\
 & \multicolumn{2}{c}{\cellcolor[HTML]{ADADAD}\textbf{Exemplar Size}} & \multicolumn{4}{c}{\textbf{100}} & \multicolumn{4}{c}{\textbf{394}} & \multicolumn{4}{c}{\textbf{1550}} & \multicolumn{4}{c}{\textbf{6100}} & \multicolumn{4}{c}{\textbf{60/task (eq. VAE)}} & \cellcolor[HTML]{ADADAD}\textbf{} & \cellcolor[HTML]{ADADAD}\textbf{} \\
 \multicolumn{1}{|c|}{} & \multicolumn{1}{c|}{} & \textbf{2-1-1-1-1} & 0.25 & 0.24 & 0.26 & 0.28 & 0.29 & 0.30 & 0.29 & 0.31 & 0.33 & 0.32 & 0.30 & 0.33 & 0.33 & 0.34 & 0.30 & 0.33 & 0.29 & 0.29 & 0.29 & 0.31 & \textbf{0.46} & 0.45 \\
 \multicolumn{1}{|c|}{} & \multicolumn{1}{c|}{} & \textbf{3-1-1-1} & 0.25 & 0.25 & 0.27 & 0.29 & 0.29 & 0.29 & 0.30 & 0.31 & 0.32 & 0.32 & 0.31 & 0.32 & 0.34 & 0.33 & 0.30 & 0.33 & 0.28 & 0.27 & 0.30 & 0.30 & \textbf{0.46} & 0.45 \\
 \multicolumn{1}{|c|}{} & \multicolumn{1}{c|}{} & \textbf{2-2-2} & 0.40 & 0.41 & 0.39 & 0.44 & 0.44 & 0.43 & 0.43 & 0.46 & 0.45 & 0.46 & 0.44 & 0.46 & 0.46 & 0.46 & 0.44 & 0.48 & 0.42 & 0.43 & 0.41 & 0.45 & \textbf{0.65} & \textbf{0.65} \\
 \multicolumn{1}{|c|}{} & \multicolumn{1}{c|}{} & \textbf{2-3-1} & 0.27 & 0.27 & 0.29 & 0.30 & 0.30 & 0.31 & 0.32 & 0.32 & 0.33 & 0.33 & 0.32 & 0.34 & 0.34 & 0.34 & 0.30 & 0.34 & 0.27 & 0.28 & 0.31 & 0.30 & \textbf{0.54} & \textbf{0.54} \\
 \multicolumn{1}{|c|}{} & \multicolumn{1}{c|}{} & \textbf{3-2-1} & 0.26 & 0.27 & 0.30 & 0.30 & 0.30 & 0.31 & 0.30 & 0.32 & 0.32 & 0.33 & 0.31 & 0.33 & 0.33 & 0.34 & 0.31 & 0.34 & 0.28 & 0.28 & 0.30 & 0.30 & \textbf{0.49} & \textbf{0.49} \\
 \multicolumn{1}{|c|}{} & \multicolumn{1}{c|}{} & \textbf{3-3} & 0.54 & 0.54 & 0.58 & 0.59 & 0.58 & 0.57 & 0.61 & 0.60 & 0.60 & 0.59 & 0.59 & 0.60 & 0.60 & 0.60 & 0.59 & 0.61 & 0.54 & 0.56 & 0.58 & 0.59 & \textbf{0.81} & \textbf{0.81} \\
\multicolumn{1}{|c|}{\multirow{-7}{*}{\rotatebox[origin=c]{90}{\textbf{HHAR}}}} & \multicolumn{1}{c|}{\multirow{-7}{*}{\rotatebox[origin=c]{90}{\textbf{Scenario}}}} & \textbf{4-2} & 0.40 & 0.41 & 0.40 & 0.47 & 0.46 & 0.44 & 0.42 & 0.46 & 0.46 & 0.47 & 0.41 & 0.47 & 0.46 & 0.47 & 0.42 & 0.49 & 0.41 & 0.42 & 0.39 & 0.46 & \textbf{0.71} & \textbf{0.71} \\
 & \multicolumn{2}{c}{\cellcolor[HTML]{ADADAD}\textbf{Exemplar Size}} & \multicolumn{4}{c}{\textbf{100}} & \multicolumn{4}{c}{\textbf{303}} & \multicolumn{4}{c}{\textbf{921}} & \multicolumn{4}{c}{\textbf{2795}} & \multicolumn{4}{c}{\textbf{60/task (eq. VAE)}} & \cellcolor[HTML]{ADADAD}\textbf{} & \cellcolor[HTML]{ADADAD}\textbf{} \\
 \multicolumn{1}{|c|}{} & \multicolumn{1}{c|}{} & \textbf{4-1-1-1-1} & 0.28 & 0.29 & 0.26 & 0.33 & 0.31 & 0.31 & 0.26 & 0.34 & 0.33 & 0.33 & 0.26 & 0.33 & \textbf{0.35} & 0.34 & 0.25 & 0.34 & 0.31 & 0.31 & 0.26 & 0.33 & 0.33 & 0.34 \\
 \multicolumn{1}{|c|}{} & \multicolumn{1}{c|}{} & \textbf{2-3-1-1-1} & 0.28 & 0.28 & 0.25 & 0.33 & 0.31 & 0.30 & 0.26 & 0.33 & 0.33 & 0.33 & 0.25 & 0.32 & \textbf{0.34} & \textbf{0.34} & 0.25 & \textbf{0.34} & 0.31 & 0.31 & 0.26 & 0.32 & \textbf{0.34} & \textbf{0.34} \\
 \multicolumn{1}{|c|}{} & \multicolumn{1}{c|}{} & \textbf{3-2-1-1-1} & 0.28 & 0.28 & 0.27 & 0.32 & 0.31 & 0.31 & 0.26 & 0.33 & 0.34 & 0.32 & 0.25 & 0.34 & 0.34 & 0.34 & 0.25 & \textbf{0.36} & 0.31 & 0.31 & 0.26 & 0.33 & 0.34 & 0.34 \\
 \multicolumn{1}{|c|}{} & \multicolumn{1}{c|}{} & \textbf{4-2-1-1} & 0.28 & 0.29 & 0.27 & 0.33 & 0.31 & 0.31 & 0.27 & 0.34 & 0.35 & 0.34 & 0.26 & 0.35 & 0.35 & 0.35 & 0.26 & 0.36 & 0.31 & 0.31 & 0.27 & 0.34 & 0.40 & \textbf{0.41} \\
 \multicolumn{1}{|c|}{} & \multicolumn{1}{c|}{} & \textbf{2-2-2-2} & 0.37 & 0.38 & 0.36 & 0.44 & 0.43 & 0.43 & 0.36 & 0.46 & 0.45 & 0.44 & 0.37 & 0.47 & 0.46 & 0.45 & 0.36 & 0.45 & 0.42 & 0.42 & 0.36 & 0.45 & \textbf{0.53} & \textbf{0.53} \\
 \multicolumn{1}{|c|}{} & \multicolumn{1}{c|}{} & \textbf{3-3-2} & 0.41 & 0.41 & 0.38 & 0.44 & 0.44 & 0.44 & 0.39 & 0.46 & 0.46 & 0.47 & 0.38 & 0.48 & 0.45 & 0.46 & 0.37 & 0.47 & 0.41 & 0.42 & 0.38 & 0.45 & 0.58 & \textbf{0.59} \\
 \multicolumn{1}{|c|}{} & \multicolumn{1}{c|}{} & \textbf{4-2-2} & 0.39 & 0.39 & 0.37 & 0.41 & 0.42 & 0.42 & 0.38 & 0.44 & 0.46 & 0.46 & 0.38 & 0.46 & 0.46 & 0.45 & 0.37 & 0.46 & 0.42 & 0.41 & 0.37 & 0.43 & \textbf{0.59} & \textbf{0.59} \\
\multicolumn{1}{|c|}{\multirow{-6}{*}{\rotatebox[origin=c]{90}{\textbf{RealWorld}}}} & \multicolumn{1}{c|}{\multirow{-6}{*}{\rotatebox[origin=c]{90}{\textbf{Scenario}}}} & \textbf{5-3} & 0.50 & 0.50 & 0.44 & 0.59 & 0.55 & 0.54 & 0.43 & 0.58 & 0.57 & 0.58 & 0.39 & 0.58 & 0.59 & 0.59 & 0.36 & 0.56 & 0.52 & 0.51 & 0.45 & 0.57 & \textbf{0.73} & \textbf{0.73} \\
 & \multicolumn{2}{c}{\cellcolor[HTML]{ADADAD}\textbf{Exemplar Size}} & \multicolumn{4}{c}{\textbf{100}} & \multicolumn{4}{c}{\textbf{240}} & \multicolumn{4}{c}{\textbf{575}} & \multicolumn{4}{c}{\textbf{1378}} & \multicolumn{4}{c}{\textbf{60/task (eq. VAE)}} & \cellcolor[HTML]{ADADAD}\textbf{} & \cellcolor[HTML]{ADADAD}\textbf{} \\
 \multicolumn{1}{|c|}{} & \multicolumn{1}{c|}{} & \textbf{5-1-1-1-1-1} & 0.20 & 0.20 & 0.21 & 0.21 & 0.22 & 0.22 & 0.19 & 0.20 & 0.22 & 0.23 & 0.14 & 0.20 & 0.21 & 0.22 & 0.14 & 0.19 & 0.23 & 0.23 & 0.16 & 0.20 & 0.30 & \textbf{0.31} \\
 \multicolumn{1}{|c|}{} & \multicolumn{1}{c|}{} & \textbf{2-2-2-2-2} & 0.28 & 0.28 & 0.25 & 0.29 & 0.29 & 0.29 & 0.19 & 0.27 & 0.30 & 0.30 & 0.16 & 0.28 & 0.28 & 0.30 & 0.18 & 0.27 & 0.29 & 0.29 & 0.18 & 0.27 & \textbf{0.45} & \textbf{0.45} \\
 \multicolumn{1}{|c|}{} & \multicolumn{1}{c|}{} & \textbf{4-3-2-1} & 0.21 & 0.21 & 0.21 & 0.22 & 0.22 & 0.23 & 0.20 & 0.21 & 0.23 & 0.23 & 0.15 & 0.21 & 0.23 & 0.23 & 0.13 & 0.22 & 0.22 & 0.23 & 0.20 & 0.21 & 0.40 & \textbf{0.41} \\
 \multicolumn{1}{|c|}{} & \multicolumn{1}{c|}{} & \textbf{2-3-4-1} & 0.21 & 0.21 & 0.20 & 0.23 & 0.23 & 0.22 & 0.19 & 0.23 & 0.24 & 0.24 & 0.16 & 0.22 & 0.23 & 0.23 & 0.15 & 0.21 & 0.23 & 0.22 & 0.19 & 0.23 & \textbf{0.47} & \textbf{0.47} \\
 \multicolumn{1}{|c|}{} & \multicolumn{1}{c|}{} & \textbf{4-3-3} & 0.36 & 0.36 & 0.21 & 0.32 & 0.37 & 0.37 & 0.18 & 0.33 & 0.38 & 0.37 & 0.15 & 0.34 & 0.37 & 0.36 & 0.15 & 0.31 & 0.37 & 0.37 & 0.18 & 0.33 & \textbf{0.60} & \textbf{0.60} \\
 \multicolumn{1}{|c|}{} & \multicolumn{1}{c|}{} & \textbf{5-3-2} & 0.29 & 0.29 & 0.23 & 0.30 & 0.29 & 0.31 & 0.17 & 0.28 & 0.30 & 0.30 & 0.15 & 0.28 & 0.31 & 0.31 & 0.15 & 0.27 & 0.29 & 0.29 & 0.18 & 0.27 & 0.54 & \textbf{0.55} \\
\multicolumn{1}{|c|}{\multirow{-6}{*}{\rotatebox[origin=c]{90}{\textbf{PAMAP2}}}} & \multicolumn{1}{c|}{\multirow{-6}{*}{\rotatebox[origin=c]{90}{\textbf{Scenario}}}} & \textbf{6-4} & 0.43 & 0.44 & 0.19 & 0.39 & 0.47 & 0.47 & 0.16 & 0.38 & 0.46 & 0.46 & 0.14 & 0.35 & 0.46 & 0.46 & 0.13 & 0.36 & 0.44 & 0.45 & 0.19 & 0.41 & \textbf{0.72} & \textbf{0.72}
\end{tabular}}
\end{table*}

\paragraph*{Plasticity of the CL classifier}
The new-class accuracy for each method is presented in Fig. \ref{fig:new_acc}. All methods show similar performance when dealing with larger datasets (bottom of the figure), indicating their ability to generalize to new classes as the amount of training data increases. However, when working with smaller datasets (top of the figure), iCaRL and LUCIR exhibit a noticeable decline in performance.

\paragraph*{Stability of the CL classifier}
The old-class accuracy for each method is shown in Fig. \ref{fig:old_acc}. Despite a significant decline in the accuracy in all sampling methods from Task 2, TaskVAE maintains the highest accuracy when transitioning to new tasks. Specifically, TaskVAE achieves approximately 70\% accuracy in Task 2 and nearly 50\% in Task 3. Notably, TaskVAE's performance remains stable even as the dataset size increases.

Our findings highlight that the choice of replay strategy in the performance of CL classifiers plays a crucial role in the plasticity and stability performance of the model. Our extensive analysis indicates that TaskVAE outperforms experience replay methods via exemplars while maintaining the same memory cost. One of the key strengths of TaskVAE is its consistent performance, even when faced with smaller datasets. While the other models struggle with limited data, our approach shows better generalization capabilities. This is particularly valuable in practical applications where large datasets may not always be available.

Furthermore, TaskVAE demonstrates robust stability as the dataset size increases. Unlike the other models, our approach retains its effectiveness and continues to perform well without significant degradation in accuracy. This makes it a reliable choice for tasks requiring long-term performance across varying conditions.

In addition to its performance advantages, TaskVAE is also highly memory-efficient. The model's memory footprint is equivalent to the holdout size of just 60 samples per task, which is considerably lower than that of traditional experience replay methods. Despite this compact memory requirement, the VAE component of TaskVAE allows it to generate an unlimited number of synthetic samples as needed. This capability ensures that the model remains flexible and scalable.

\section{Conclusion and Future Work}
\label{sec:conclusion}

In conclusion, TaskVAE presents an effective solution for replay-based continual learning in class-incremental settings. By leveraging task-specific VAEs to generate synthetic exemplars, TaskVAE maintains strong knowledge retention, adaptability, and memory efficiency, outperforming traditional replay methods. Its ability to handle smaller datasets, retain performance as dataset size increases, and offer scalable memory management makes it a robust and practical choice for real-world applications, particularly in domains like Human Activity Recognition. TaskVAE’s contributions lie in its novel approach to balancing memory constraints, task-specific generation, and long-term stability, paving the way for more efficient and reliable continual learning models.

Future work on TaskVAE could explore integrating advanced generative models \cite{sohl2015deep, reynolds2009gaussian} to improve the quality of synthetic samples and better capture complex data distributions. Additionally, further optimization of the sampling and filtering mechanisms could enhance its effectiveness, ensuring more reliable generation of synthetic samples and improving stability and knowledge retention in class-incremental learning tasks.

\bibliographystyle{ieeetr}
\bibliography{reference}

\end{document}